\documentclass[pdflatex,sn-mathphys-num]{sn-jnl}

\usepackage{graphicx}%
\usepackage{multirow}%
\usepackage{amsmath,amssymb,amsfonts}%
\usepackage{amsthm}%
\usepackage{mathrsfs}%
\usepackage[title]{appendix}%
\usepackage{xcolor}%
\usepackage{textcomp}%
\usepackage{manyfoot}%
\usepackage{booktabs}%
\usepackage{algorithm}%
\usepackage{algorithmicx}%
\usepackage{algpseudocode}%
\usepackage{listings}%
\usepackage{lineno,hyperref}
\usepackage{epstopdf}
\usepackage{nicefrac}
\usepackage{makecell}
\usepackage{subcaption}

\theoremstyle{thmstyleone}%
\theoremstyle{thmstyletwo}%
\newtheorem{example}{Example}%

\theoremstyle{thmstylethree}%

\begin{document}

\title[Advanced Physics-Informed Neural Network with Residuals for Solving Complex Integral Equations]{Advanced Physics-Informed Neural Network with Residuals for Solving Complex Integral Equations}


\author[1]{\fnm{Mahdi} \sur{Movahedian Moghaddam}}\email{m\_movahedian@sbu.ac.ir}

\author*[1,2]{\fnm{Kourosh} \sur{Parand}}\email{k\_parand@sbu.ac.ir}

\author*[1]{\fnm{Saeed Reza} \sur{Kheradpisheh}}\email{s\_kheradpisheh@sbu.ac.ir}

\affil*[1]{\orgdiv{Department of Computer and Data Sciences}, \orgname{Shahid Beheshti University}, \orgaddress{
\city{Tehran},
\state{Tehran}, \country{Iran}}}

\affil[2]{\orgdiv{Department of Cognitive Modeling}, \orgname{Shahid Beheshti University}, \orgaddress{
\city{Tehran},
\state{Tehran},
\country{Iran}}}



\abstract{In this paper, we present the Residual Integral Solver Network (RISN), a novel neural network architecture designed to solve a wide range of integral and integro-differential equations, including one-dimensional, multi-dimensional, ordinary and partial integro-differential, systems, fractional types, and Helmholtz-type integral equations involving oscillatory kernels. RISN integrates residual connections with high-accuracy numerical methods such as Gaussian quadrature and fractional derivative operational matrices, enabling it to achieve higher accuracy and stability than traditional Physics-Informed Neural Networks (PINN). The residual connections help mitigate vanishing gradient issues, allowing RISN to handle deeper networks and more complex kernels, particularly in multi-dimensional problems. Through extensive experiments, we demonstrate that RISN consistently outperforms not only classical PINNs but also advanced variants such as Auxiliary PINN (A-PINN) and Self-Adaptive PINN (SA-PINN), achieving significantly lower Mean Absolute Errors (MAE) across various types of equations. These results highlight RISN’s robustness and efficiency in solving challenging integral and integro-differential problems, making it a valuable tool for real-world applications where traditional methods often struggle.
}




\maketitle

\section{Introduction}

Integral and integro-differential equations are foundational tools in many fields of science and engineering, modeling a wide range of phenomena from physics and biology to economics and engineering systems \cite{wazwaz2011linear,kress1989linear,tebeest1997classroom}. These equations describe processes that depend not only on local variables but also on historical or spatial factors, making them essential for understanding systems with memory effects, hereditary characteristics, and long-range interactions \cite{kermack1927contribution,estrada2012singular,scudo1971vito,minakov2018non}. Despite their importance, solving integral and integro-differential equations is a challenging task due to the complexity of their integral operators, especially when extended to multi-dimensional or fractional forms \cite{khuri1996decomposition,kress1989linear}.
\par
Classical numerical methods, such as finite difference \cite{thomas2013numerical,dehghan2006finite}, finite element \cite{zienkiewicz2005finite,david2020application}, and spectral methods \cite{boyd2001chebyshev,yousefi2022numerical,latifi2020generalized}, have long been used to approximate solutions to these equations. However, these methods often suffer from several limitations. For example, they require fine discretization of the problem domain, leading to high computational costs and memory usage, especially for high-dimensional problems \cite{griebel2010dimension,bastian2008generic,ciarlet2002finite,sundar2012parallel}. Additionally, these methods can struggle with non-local operators, singular kernels, or fractional derivatives, which introduce further complexities \cite{tarasov2011fractional}. Even though more recent approaches, like Gaussian quadrature and collocation methods, provide enhanced accuracy for certain cases, their applicability is often restricted to simpler, well-behaved systems \cite{khuri1996decomposition,atkinson1991introduction,estrada2012singular,tarasov2011fractional,saadatmandi2011legendre}.
\par
In recent years, deep learning has emerged as a powerful tool to overcome the limitations of traditional statistical methods \cite{goodfellow2016deep,lecun2015deep}. Physics-informed neural networks (PINNs) represent a major advance in this direction by directly integrating the governing physical laws with loss of neural activity \cite{raissi2019physics,karniadakis2021physics,lu2021deepxde}. This allows PINNs to learn solutions to differential equations using observational data while ensuring consistency with underlying physics \cite{raissi2018hidden,zhu2019data}. PINNs have shown promising results in solving various types of differential equations, including ordinary and partial differential equations \cite{berg2018unified,sirignano2018dgm,han2018solving}. However, PINNs are not without their limitations. The depth of the network often exacerbates issues such as vanishing gradients, and the absence of residual connections can make training deep networks unstable, especially when dealing with highly complex or multi-dimensional integral and integro-differential equations \cite{he2016deep,bengio1994learning,hochreiter1998vanishing,gazoulis2023stability}. Although extensions like Auxiliary PINN (A-PINN) \cite{yuan2022pinn} and Self-Adaptive PINN (SA-PINN) \cite{mcclenny2023self} have been developed to address some limitations, they still encounter significant challenges in solving complex integral and integro-differential equations, motivating the need for the RISN architecture.
\par
To overcome these challenges, we propose the Residual Integral Solver Network (RISN), a novel extension of the PINN framework specifically designed to solve integral and integro-differential equations. RISN enhances the original PINN architecture by incorporating residual connections, a proven technique in deep learning that significantly improves gradient flow, reduces training instability, and allows for deeper network architectures. These residual connections are crucial for ensuring stable training, particularly when handling multi-dimensional or highly non-linear systems \cite{he2016deep,raissi2019physics,han2018solving}. Furthermore, RISN leverages high-accuracy numerical techniques, such as Gaussian quadrature for efficient and precise calculation of integral terms, and fractional operational matrices to handle fractional derivatives with minimal error \cite{atkinson1991introduction,aghaei2024pinniesefficientphysicsinformedneural}. This combination of deep learning with classical numerical methods allows RISN to achieve high accuracy and stability across a wide range of equation types, including those that pose challenges for traditional methods and standard PINNs.
\par
In this paper, we contribute to the growing field of deep learning for solving complex mathematical problems by presenting the following innovations: (1) the development of a residual-based neural network model capable of solving integral and integro-differential equations of various types, including one-dimensional, multi-dimensional, fractional systems, and oscillatory Helmholtz-type integral equations; (2) the integration of advanced numerical techniques within the RISN framework to enhance precision in the calculation of integral and differential operators; and (3) an extensive evaluation of the RISN model across a broad spectrum of integral equations, demonstrating superior performance in terms of accuracy, stability, and efficiency when compared to classical PINNs.
\par
The rest of this paper is organized as follows: Section 2 provides a literature review of existing methods for solving integral and integro-differential equations, with a focus on the strengths and limitations of classical numerical approaches and PINNs. In Section 3, we introduce the RISN model in detail, including its architecture, the integration of residual connections, and the numerical techniques employed. Section 4 presents the experimental results, showcasing the RISN model's performance across different equation types. Finally, in Section 5, we discuss the implications of our findings and propose directions for future work.

\section{Literature Review}
The solution of integral and integro-differential equations plays a critical role across various scientific disciplines, including physics, biology, and engineering. These equations model systems with local and non-local interactions, which makes them essential for understanding phenomena ranging from heat conduction and population dynamics to quantum mechanics and fluid dynamics. Traditionally, methods such as finite difference and finite element approaches have been employed to solve these equations \cite{wang2008finite,brambilla2008integral}. Although these techniques have been widely successful, they face significant computational challenges, particularly when applied to high-dimensional problems \cite{vandandoo2024high,Hamfeldt2021,Sarakorn2018}.
\par
The finite difference method (FDM) approximates differential operators by discretizing the domain, making it simple to implement for many problems. However, FDM struggles with large-scale problems due to the need for fine grid discretization, which leads to high computational cost and memory usage \cite{Liu2022}. Finite element methods (FEM), while more versatile and effective for irregular domains, encounter similar issues in high-dimensional systems, where the complexity of mesh generation and element assembly increases exponentially with the number of dimensions \cite{Nath2024,Liu2022}. The inefficiency of these methods in handling complex multi-dimensional and fractional equations has spurred the development of alternative approaches, including machine learning-based methods \cite{math11214418,Nath2024}.
\par
In recent years, Artificial Neural Networks (ANNs) have emerged as a powerful alternative for solving complex mathematical equations, including Fredholm and Volterra integral equations. Early work by \cite{effati2012neural} demonstrated the use of ANNs in solving Fredholm integral equations, highlighting their potential for improving computational efficiency and accuracy. Further advancements were made by \cite{mosleh2014numerical}, who extended neural network methods to handle fuzzy integro-differential equations, paving the way for the application of ANNs to more complex and uncertain systems.
\par
As research into ANNs progressed, attention shifted to solving more complex problems, such as multi-dimensional and fractional integro-differential equations. The research \cite{jafarian2015artificial} successfully applied ANNs to Volterra integral equations, demonstrating superior performance compared to traditional methods in terms of both speed and accuracy. In parallel, \cite{asady2014utilizing} employed ANNs to tackle two-dimensional integral equations, while \cite{chaharborj2017study} utilized ANNs for fractional-order equations, both demonstrating significant improvements in computational efficiency.
\par
A major leap forward in this field came with the introduction of Physics-Informed Neural Networks (PINNs). PINNs embed the physical laws governing a system directly into the neural network’s architecture, allowing for the solution of complex partial differential equations (PDEs) and integro-differential equations with increased accuracy. For instance, \cite{guo2021physics} demonstrated how PINNs could be used to solve volume integral equations, combining physical insights with deep learning techniques to ensure physically consistent solutions. Furthermore, the DeepXDE framework, introduced by \cite{lu2021deepxde}, extended PINNs to a wide variety of differential and integro-differential equations, emphasizing their flexibility in handling real-world problems.
\par
Recent approaches, such as DeepONets \cite{Lu2021} and Fourier Neural Operators \cite{li2021fourierneuraloperatorparametric}, have pushed the boundaries of solving high-dimensional equations. However, these models tend to struggle with generalization across different types of equations, particularly those involving non-local operators or fractional components.
\par
An important development in this area is the PINNIES framework \cite{aghaei2024pinniesefficientphysicsinformedneural}, which integrates physics-informed strategies to handle integral operator problems. This framework has been particularly effective in addressing multi-dimensional and fractional equations, offering enhanced accuracy and stability compared to traditional methods.
\par
However, despite their successes, deep neural networks, including PINNs, suffer from the vanishing gradient problem, which occurs when the gradients used for backpropagation become very small in deep architectures. This results in slow or stalled learning as the network struggles to adjust its weights. To mitigate this issue, residual connections were introduced in deep learning models. These connections, first proposed by \cite{he2016deep} in the context of ResNets, allow gradients to flow more easily through deeper layers, making training more stable and efficient. Recent adaptations of residual learning into the PINN framework have been shown to effectively handle high-dimensional problems and integral equations, improving both convergence speed and solution accuracy.
\par
In addition to PINNs, recent works have introduced innovative neural network architectures for solving fractional-order equations and integro-differential systems. For example, \cite{martire2022fractional} applied ANNs to fractional Volterra integral equations, while \cite{Saneifard_Jafarian_Ghalami_Nia_2022} explored extended neural networks for solving two-dimensional fractional equations, achieving superior precision through techniques such as sine-cosine basis functions and extreme learning machines (ELM). These methods provide robust solutions to problems that classical approaches often find intractable.
\par
Several variants of PINNs have been proposed to improve stability and convergence, such as the Auxiliary PINN (A-PINN) \cite{yuan2022pinn}, which augments the loss function with auxiliary conditions, and the Self-Adaptive PINN (SA-PINN) \cite{mcclenny2023self}, which introduces adaptive weighting schemes for the loss components. While these methods offer improvements in specific contexts, they often fail to generalize to more complex, multi-dimensional, and fractional problems due to the lack of structural enhancements like residual connections
\par
Furthermore, the evolution of neural network techniques, particularly the development of PINNs (such as PINNIES, and A-PINN), has transformed the landscape of integral equation solving. These frameworks offer significant improvements in computational efficiency, accuracy, and the ability to handle complex, high-dimensional, and fractional systems, representing a substantial advancement over traditional methods.

\section{Methodology}\label{sec:3}
The Residual Integral Solver Network (RISN) is a novel neural network architecture to address integral and integro-differential equations efficiently. Inspired by traditional neural networks, RISN integrates residual connections to mitigate the vanishing gradient problem and enable deeper network training. The model is structured to handle both differential and integral operators by incorporating these connections, which not only improve training stability but also enhance the accuracy of the solutions. By combining classical numerical techniques like Gaussian numerical quadrature and fractional operational matrices with the residual structure, RISN is capable of solving a wide range of complex mathematical problems, including integral, fractional integro-differential, and oscillatory Helmholtz-type integral equations. This structure is chosen for solving integral equations due to its ability to maintain stability and accuracy in deep networks. This section provides a detailed explanation of the RISN structure and the numerical techniques employed to achieve high precision in solving these equations.

\subsection{The Model Structure}
The RISN is an advanced neural network architecture designed to address integral and integro-differential equations by integrating residual connections. The structure of RISN is inspired by traditional neural networks but enhances them with residual connections to improve training efficiency and solution accuracy.
\par
Residual connections, a critical component in our model, allow for the preservation of gradient flow during backpropagation, thereby preventing the vanishing gradient problem typically seen in deep networks. This feature is particularly important in multi-dimensional and fractional integro-differential equations, where the complexity of the solution requires a stable and accurate gradient propagation mechanism.
\begin{figure}
    \centering
    \includegraphics[width=\linewidth]{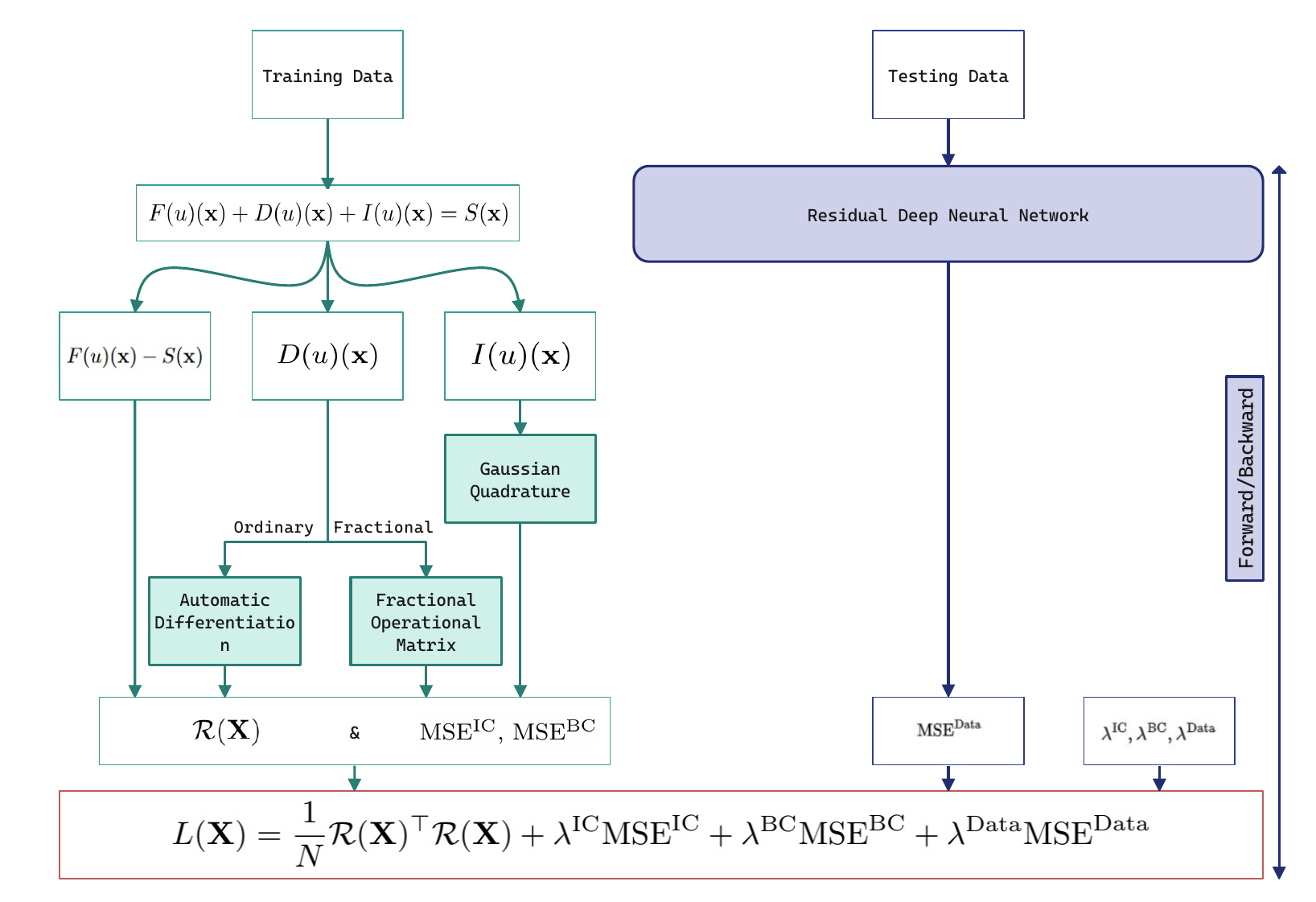}
    \caption{Flowchart of the RISN methodology. The equation \( F(u)(\mathbf{x}) + D(u)(\mathbf{x}) + I(u)(\mathbf{x}) = S(\mathbf{x}) \) is decomposed into three components: \(F(u)(\mathbf{x}) + S(\mathbf{x})\), \(D(u)(\mathbf{x})\), and \(I(u)(\mathbf{x})\). The fractional derivative terms are handled using fractional operational matrices, while Gaussian quadrature is employed for the numerical quadrature of the integral term \( I(u)(\mathbf{x}) \). The output is optimized through residual deep neural networks, where mean squared error (MSE) terms for initial conditions, boundary conditions, and data are combined to form the loss function \( L(x) \), ensuring accurate predictions.}
    \label{fig:risn}
\end{figure}
\par
The architecture of the RISN, as illustrated in Figure \ref{fig:risn}, is designed to solve integral and integro-differential equations by combining residual deep neural networks with classical numerical methods. This figure provides an overview of the data flow and interaction between different parts of the network. The equation (\ref{eq:ie}) represents the target system, where \( F(u)(\mathbf{x}) \), \( D(u)(\mathbf{x}) \), and \( I(u)(\mathbf{x}) \) correspond to the differential, integral, and source terms, respectively. 
\begin{equation}
    F(u)(\mathbf{x}) + D(u)(\mathbf{x}) + I(u)(\mathbf{x}) = S(\mathbf{x}),
    \label{eq:ie}
\end{equation}
\par
The model begins with the \textit{training data}, which feeds into different branches corresponding to the components of the equation.
The source and function terms are handled through the combination of \( F(u)(\mathbf{x}) + S(\mathbf{x}) \).
For the differential operator \( D(u)(\mathbf{x}) \), two approaches are used: \textit{automatic differentiation} for ordinary derivatives and the \textit{fractional operational matrix} for fractional derivatives.
The integral term \( L(\mathbf{X}) \) is computed using \textit{Gaussian quadrature}, a highly accurate method for numerical quadrature.
\par
Meanwhile, \textit{testing data} is passed through the residual deep neural network. This neural network incorporates residual connections, which enhance training stability and enable the model to learn deeper networks without suffering from the vanishing gradient problem. This allows the RISN to converge more effectively and improve accuracy in solving complex integral and integro-differential equations.
\par
The residual network model that should find $u(x)$ is defined as follows:
\begin{equation}
\begin{aligned}
    A_0 &= \mathbf{X}, & \mathbf{X} \in \mathbb{R}^{N\times d}, \\
    A_i &= \sigma_i(A_{i-1}\boldsymbol{\theta}^{(i)}+\mathbf{b}^{(i)}) + A_{i-1}, & i = 1,2,\ldots L-1,\\
    \mathrm{MLP}(\mathbf{X}) &:= A_L = A_{L-1}\boldsymbol{\theta}^{(L)}+\mathbf{b}^{(L)}, & A_L \in \mathbb{R}^{N\times 1}.
\end{aligned}    
\end{equation}
The input to the model is denoted by $\mathbf{X}$, which is a matrix with $N$ data points, each having $d$ dimensions. For $i = 1,2,\ldots L-1$, each hidden layer $A_i$  is computed by applying a nonlinear activation function $\sigma_i$ to a linear transformation of the previous layer $A_{i-1}$ with parameters $\boldsymbol{\theta}^{(i)}$ and bias $\mathbf{b}^{(i)}$, and then adding the residual connection from the previous layer $A_{i-1}$. This residual connection helps in mitigating the vanishing gradient problem and allows for deeper network training. The final output $A_L$  is produced by applying a linear transformation to the last hidden layer $A_{L-1}$ with parameters $\boldsymbol{\theta}^{(L)}$ and bias $\mathbf{b}^{(L)}$.
\par
The objective is to train the RISN to minimize the residual:
\begin{equation*}
    \mathcal{R}(\mathbf{X}) := F(\mathbf{u}) + D(\mathbf{u}) + I(\mathbf{u}) - S(\mathbf{X}).
\end{equation*}

After processing through both the neural network and numerical techniques, the model computes several types of error, including \textit{mean squared error (MSE)} for data-driven predictions and separate errors for initial conditions (MSE\(^\text{IC}\)) and boundary conditions (MSE\(^\text{BC}\)). These errors are combined into the overall loss function (\ref{eq:loss}), which also includes the residual terms \( \mathcal{R}(x) \). The parameters $\lambda^{\text{IC}}$, $\lambda^{\text{IC}}$, and $\lambda^{\text{Data}}$ serve as regularization coefficients. The final loss function is a weighted sum of the errors, using parameters \( \lambda \) to balance the contributions from each component. 
\begin{equation}
    L(\mathbf{X}) = \frac{1}{N} \mathcal{R}(\mathbf{X})^\top \mathcal{R}(\mathbf{X}) + \lambda^{\text{IC}} \mathrm{MSE}^{\text{IC}} + \lambda^{\text{BC}} \mathrm{MSE}^{\text{BC}} + \lambda^{\text{Data}} \mathrm{MSE}^{\text{Data}},
    \label{eq:loss}
\end{equation}
When solving a system of \(M\) equations, the loss function \(L(\mathbf{X})\) is computed as the sum of residuals across all equations, normalized by the product of the number of training data points \(N\) and the number of equations \(M\). The general form of the loss function is given by:

\begin{equation}
L(\mathbf{X}) = \frac{1}{N \times M} \sum_{\iota=1}^M \mathcal{R}_\iota(\mathbf{X})^\top \mathcal{R}_\iota(\mathbf{X}) + \sum_{\iota=1}^M \left( \lambda_\iota^{\text{IC}} \mathrm{MSE}_\iota^{\text{IC}} + \lambda_\iota^{\text{BC}} \mathrm{MSE}_\iota^{\text{BC}} + \lambda_\iota^{\text{Data}} \mathrm{MSE}_\iota^{\text{Data}} \right).    
\end{equation}

Here, \( \mathcal{R}_\iota(\mathbf{X}) \) represents the residual of the \(\iota\)-th equation, and the mean squared error terms \( \mathrm{MSE}_\iota^{\text{IC}}, \mathrm{MSE}_\iota^{\text{BC}}, \mathrm{MSE}_\iota^{\text{Data}} \) correspond to the initial condition (IC), boundary condition (BC), and testing data, respectively. The weighting factors \( \lambda_\iota^{\text{IC}}, \lambda_\iota^{\text{BC}}, \lambda_\iota^{\text{Data}} \) control the relative contribution of each term to the overall loss, ensuring that the optimization process takes into account the specific requirements of the initial, boundary, and data constraints for each equation in the system.

\subsection{Numerical Techniques for Integral and Fractional Differential Operators}
In this work, two key numerical techniques have been employed to enhance the accuracy and efficiency of solving integral and fractional integro-differential equations: Gaussian numerical quadrature and fractional operational matrices. Both methods, though well-established in the field of numerical analysis, were first integrated into the PINN framework in the work of \cite{aghaei2024pinniesefficientphysicsinformedneural}. The use of these techniques enhances numerical accuracy and reduces errors from numerical approximations. Their integration into the RISN further solidifies the model’s capacity to tackle complex mathematical problems.

\subsubsection{Gaussian Numerical Quadrature}
Gaussian quadrature, particularly well-suited for evaluating definite integrals, is leveraged in this model to compute the integral terms that arise in the solution of integro-differential equations. As described in the work of PINNIES, this method offers a high degree of accuracy with relatively few evaluation points, making it a computationally efficient choice. In the context of RISN, Gaussian quadrature allows for the precise calculation of integral terms, which are critical to solving the targeted problems. Its inclusion ensures that the integral operators are handled with minimal numerical error, thus enhancing the overall accuracy of the model.

\subsubsection{Fractional Operational Matrices}
Fractional differential operators are essential when dealing with integro-differential equations that exhibit non-local behavior, such as systems with memory effects. The method of using fractional operational matrices, initially introduced in the context of PINNs by PINNIES, allows for a structured and efficient way of handling these operators. By representing fractional derivatives in matrix form, the RISN model can compute fractional derivatives with high precision, seamlessly integrating this approach into the neural network architecture. This not only improves the accuracy of fractional derivative calculations but also streamlines the process within the network’s training and inference stages.
\par
In these problems, fractional derivatives, which generalize the concept of integer-order differentiation, allow for a more accurate representation of processes with memory effects or non-local dependencies. RISN, combined with operational matrices, efficiently handles these derivatives by leveraging its residual connections to maintain stability in training.
\par
The combination of these two techniques—Gaussian numerical quadrature and fractional operational matrices—within RISN reflects the practical benefits of merging classical numerical methods with advanced machine learning models. While these techniques do not significantly differ from their original implementation in the PINN framework, their application within RISN ensures that the model can solve a wide range of integral and integro-differential equations more effectively.
\par
The RISN model leverages residual connections to enhance the training process and accuracy of solving integral and integro-differential equations. By integrating residual connections with a structured loss function, RISN improves solution fidelity and robustness, making it a powerful tool for addressing complex mathematical problems. However, the proposed method may require additional tuning when dealing with highly complex problems.

\subsection{Functional Approximation Perspective in Sobolev Spaces}
To better understand the mathematical advantage of the residual formulation adopted in RISN, we analyze the problem from the viewpoint of functional approximation in Sobolev spaces.

Let $y^* \in W^{k,p}(\Omega)$ denote the exact solution of the integral equation, where $k \geq 1$, $p \geq 1$, and $\Omega \subset \mathbb{R}^d$ is a compact domain. Let $g(x)$ be a known approximation such that $\| y^* - g \|_{W^{k,p}}$ is small. Then, the residual network is tasked with approximating the correction term:

\[
r(x) := y^*(x) - g(x).
\]

Since $r$ often exhibits smoother behavior than $y^*$ itself, the approximation of $r$ using a neural network $N(x; \theta)$ becomes more efficient.

\paragraph{Sobolev Approximation Advantage.}
Using classical results on Sobolev approximation (e.g., \cite{hornik1990universal,pinkus1999approximation,mhaskar1996neural}), we know that for neural networks with sufficient width and depth, there exists a parameter $\theta$ such that:

\[
\| r - N(\cdot;\theta) \|_{W^{k,p}} \leq \varepsilon,
\]

with $\varepsilon$ decreasing at a rate depending on the smoothness of $r$ and the architecture of the network.

\paragraph{Implication.}
Thus, the total approximation error satisfies:

\[
\| y^* - \hat{y} \|_{W^{k,p}} = \| y^* - (g + N(\cdot;\theta)) \|_{W^{k,p}} = \| r - N(\cdot;\theta) \|_{W^{k,p}} \leq \varepsilon,
\]

which is potentially smaller than directly approximating $y^*$ by a network alone, particularly when $g$ captures the coarse structure of $y^*$.

This analysis shows that the residual architecture not only simplifies the learning task but also benefits from improved theoretical approximation bounds in Sobolev spaces.
\section{Experimental Results}
In this section, we present the results obtained from solving various types of integral and integro-differential equations using the RISN. The examples provided cover a range of equations, including one-dimensional and multi-dimensional problems, systems of equations, and Helmholtz-type integral equations. The structure of this section is as follows: first, we outline the setup for each example, including the equation type, domain, and kernel functions. Next, we present the results for each test case and compare the performance of RISN with classical PINN as well as advanced variants including A-PINN and SA-PINN.
\par
For all models evaluated in this study, the loss function is composed of the physics-informed residual loss and the boundary/initial condition loss. Specifically, the total loss is defined as:
\begin{equation*}
    \mathcal{L}_{total} = \mathcal{L}_{physics} + \mathcal{L}_{boundary/initial} 
\end{equation*}
where $\mathcal{L}_{physics}$ measures the violation of the underlying integral or integro-differential equation across collocation points, and $\mathcal{L}_{boundary/initial}$ enforces the satisfaction of the given boundary or initial conditions. No adaptive loss weighting strategies were applied; all components contribute equally to the total loss. This structure is consistently maintained across both the baseline PINN and the proposed RISN models to ensure a fair and direct comparison. Note that the additional residual terms in the RISN model are seamlessly integrated into the physics loss without altering the fair comparison setup.
\par
The RISN model used in these experiments consists of a fully connected neural network with seven hidden layers, each containing 20 neurons. The $Tanh$ activation function is applied in each layer, and the model is optimized using the L-BFGS optimizer with a learning rate of 0.01. The training points size and Gaussian quadrature order are set to 50, and the loss function incorporates residuals for the target equation, as well as terms for initial and boundary conditions, with appropriate weighting parameters $\lambda$. Additionally, Gaussian quadrature (specifically Gauss–Legendre) is employed for accurate numerical quadrature, and fractional operational matrices are used for handling fractional derivatives where necessary.
\par
In order to ensure a fair comparison, all models were trained using the same network architecture, with identical numbers of hidden layers and neurons per layer. Moreover, the same optimization settings, learning rates, and stopping criteria were applied across all experiments.
\par
The proposed method was implemented in Python 3.11 utilizing the PyTorch framework (Version 2.3.1), which provides automatic differentiation capabilities and supports efficient large-scale computations. All experiments were conducted on a personal computer equipped with an Intel Core i5-13420H CPU and 8 GB of RAM, operating under Microsoft Windows 11. This computational setup was sufficient to perform the training and evaluation of the RISN model across all considered problems, without encountering significant memory or performance bottlenecks.
\par
The performance of the RISN model was evaluated using the Mean Absolute Error (MAE) as the primary error metric, ensuring a clear assessment of the model’s accuracy in solving the various types of integral and integro-differential equations.
\subsection{Integral Equations}\label{sec:res_ie}
\subsubsection{One-Dimensional Integral Equations}
In this section, we address one-dimensional integral equations of the form:
\begin{equation*}
    \kappa u(x) = S(x) + \int_\Delta K(x,t) \zeta(u(t)) \, dt,
\end{equation*}
where $S(x)$ is the source term, $K(x,t)$ represents the kernel of the integral operator, and $\zeta(x)$ characterizes the linearity of the problem. The constant parameter $\kappa \in \mathbb{R}$, if $\kappa = 0 $ is a first-kind integral equation (IE); otherwise, it is classified as a second-kind IE. The domain of the problem, denoted as $\Delta = [a,b]$, applies to all equation types. For the Fredholm operator, we assume the interval $a,b = 0,1$, and for the Volterra operator, we take $g(x), h(x) = 0, x$ for $x \in \Delta$. In the case of the Volterra-Fredholm problem, the equation is given by:
\begin{equation*}
    \kappa u(x) = S(x) + \int_a^b K_f(x,t) \zeta(u(t)) \, dt + \int_0^x K_v(x,t) \zeta(u(t)) \, dt.
\end{equation*}

Using the approach outlined in Section \ref{sec:3}, we simulate a range of integral equations and provide the problem details along with the MAE results in Table \ref{tbl:1d}. It was found that, with the exception of Abel-type singular integral equations, the method yielded highly accurate results. For Abel problems, accuracy can be improved by increasing the number of nodal points or utilizing a different Gaussian quadrature method, such as an alternative to Gauss-Legendre.
\par
The table presents a comparative analysis of the MAE between the proposed RISN method and the classical PINN approach for solving one-dimensional integral equations. The results clearly indicate that the RISN method consistently achieves lower MAE across different types of integral equations, including Volterra, Volterra-Fredholm, and Abel types. This demonstrates the superior accuracy and robustness of the RISN approach, particularly in handling complex integral equations with varying kernels and source terms. Notably, for the Abel-type equations, although the errors are generally higher, the RISN still outperforms the traditional PINN, illustrating its effectiveness in tackling singular kernel challenges. As shown in the last row of the table, the traditional PINN model struggles with nonlinear Abel-type problems and is unable to solve them, whereas the proposed RISN model successfully solves them with satisfactory accuracy.

\begin{table}
\centering
\resizebox{\textwidth}{!}{%
\begin{tabular}{@{}lllllllll@{}}
\toprule
Type & $\Delta$& $\zeta(x)$ & $\kappa$ & Exact & Source term & Kernel & PINN MAE & RISN MAE \\ \midrule
Volterra & $[0,1]$ & $x$ & $1$ & $x+e^x$ & $2e^x - 1 + \frac{x^3}{6}$ & $t-x$ & $9.53\times 10^{-3}$ & $8.46\times 10^{-6}$ \\
Volterra & $[0,1]$ & $x^2$ & $1$ & $e^x$ & $e^x - \frac{1}{2}\left(e^{2x} - 1\right)$ & $1$ & $2.80\times 10^{-4}$ & $1.05\times 10^{-5}$ \\
Volterra-Fredholm & $[0,1]$ & $x$ & $1$ & $x+e^x$ & $2e^x - \frac{x}{2} - \frac{7}{3} + \frac{x^3}{6} + x e$ & $K_f = K_v = t-x$ & $1.95\times 10^{-4}$& $7.05\times 10^{-6}$ \\
Volterra-Fredholm & $[0,1]$ & $x$ & $1$ & $xe^x$ & $e^x - 1 - x$ & $K_f = x, K_v = 1$ & $3.40\times 10^{-4}$ & $1.48\times 10^{-5}$ \\
Abel & $[0,1]$ & $x$ & $0$ & $x$ & $\frac{4}{3}x^{\frac{3}{2}}$ & $\frac{-1}{\sqrt{x-t}}$ & $3.27\times 10^{-3}$ & $3.27\times 10^{-3}$ \\
Abel & $[0,1]$ & $x^3$ & $0$ & $x$ & $\frac{32}{35}x^{\frac{7}{2}}$ & $\frac{-1}{\sqrt{x-t}}$ & $-$&  $1.70\times 10^{-3}$ \\
\bottomrule
\end{tabular}}
\caption{Comparative analysis of MAE for one-dimensional integral equations using the proposed RISN versus classical PINN methods. The RISN consistently achieves lower MAE, highlighting its enhanced accuracy and reliability across various equation types and conditions.}
\label{tbl:1d}
\end{table}
The results obtained for one-dimensional Volterra and Fredholm integral equations indicate that RISN consistently achieves lower MAE compared to the baseline PINN model. The superior accuracy of RISN can be attributed to its residual connections, which enhance gradient flow and facilitate the training process, particularly in handling non-smooth or weakly singular kernels. The PINN model, by contrast, shows difficulties in precisely approximating the solution when the kernel complexity increases.

\subsubsection{Multi-Dimensional Integral Equations}
In this section, we assess the proposed approach for solving multi-dimensional integral equations. We specifically examine a two-dimensional integral equation structured as follows:
\begin{equation*}
    \kappa u(x,y) = S(x,y) + \int_{\Delta_y}\int_{\Delta_x} K(x,y,s,t) u(s,t) \, dt \, ds,
\end{equation*}
and a three-dimensional integral equation:
\begin{equation*}
    \kappa u(x,y,z) = S(x,y,z) + \int_{\Delta_z}\int_{\Delta_y}\int_{\Delta_x} K(x,y,z,r,s,t) u(r,s,t) \, dt \, ds \, dr.
\end{equation*}
These integral equations are categorized into Fredholm or Volterra types and further classified as first-kind or second-kind. Utilizing the formulations detailed in Section \ref{sec:3}, we conduct simulations on various integral equations, employing different kernel functions and exact solutions.
\par
The results in Table \ref{tbl:nd} demonstrate that RISN achieves significantly lower MAE in both Fredholm and Volterra types of equations. This indicates that RISN provides more accurate solutions, particularly in complex multi-dimensional kernel scenarios. The improvements in accuracy with RISN highlight its robustness and efficiency over traditional PINN methods.
\par
The residual connections in the RISN model have shown significant advantages, particularly in solving multi-dimensional integro-differential equations. These connections allow for better gradient flow throughout the network, reducing the likelihood of vanishing gradients and improving the model's ability to capture complex interactions between variables. This is especially important in multi-dimensional problems, where the complexity of the solution space can challenge traditional methods like PINN. By incorporating residual connections, RISN not only enhances stability but also achieves higher accuracy in these challenging scenarios, as demonstrated by the consistently lower MAE values observed in the experiments.

\begin{table}
\centering
\resizebox{\textwidth}{!}{%
\begin{tabular}{@{}lllllll@{}}
\toprule
Type & $\Delta$ & Exact& Source Term& Kernel& PINN MAE & RISN MAE \\ \midrule
Fredholm & $[0,1] \times [0,2]$ & $x^2y$ & $x^2y + \frac{4}{9}x$ & $-\frac{1}{2}xt$ & $2.85 \times 10^{-3}$ & $1.31 \times 10^{-4}$ \\

Volterra & $[0,1] \times [0,2]$ & $x+y$ & \makecell{$(x + y - 2) e^{2x + 2y} + (2 - y) e^{x + 2y} +$ \\ $(2 - x) e^{2x + y} + x + y - 2 e^{x + y}$} & $e^{x+y+s+t}$ & $4.34 \times 10^{-2}$ & $5.64 \times 10^{-3}$\\
\bottomrule
\end{tabular}}
\caption{Comparison of MAE for multi-dimensional integral equations using RISN versus classical PINN methods. RISN shows improved accuracy, demonstrating its effectiveness in handling complex Fredholm and Volterra integral equations.}
\label{tbl:nd}
\end{table}

\subsubsection{System of Integral Equations}
A system of integral equations involves multiple unknown functions that are interconnected and appear within the integral terms. For an integer \( M \geq 2 \), such a system can be expressed mathematically as:
\begin{equation}
    \kappa u_\iota(x) = S_\iota(x) + \int_\Delta \sum_{i=1}^\iota K_{\iota,i}(x,t) u_i(t) \, dt.
\end{equation}
These systems often combine different types of integral equations previously discussed. For example, when \(\kappa = 0\), the system reduces to a set of first-kind integral equations. In this section, we focus on the case where \( M = 2 \), which results in the following system of equations:

\begin{equation}
    \begin{cases}
\kappa u_{1}(x) = S_1(x) + \displaystyle\int_{\Delta} [K_{1,1}(x,t) u_1(t) + K_{1,2}(x,t) u_2(t)] \, dt, \\
\kappa u_{2}(x) = S_2(x) + \displaystyle\int_{\Delta} [K_{2,1}(x,t) u_1(t) + K_{2,2}(x,t) u_2(t)] \, dt,
\end{cases}
\end{equation}
subject to \(v\) boundary conditions for \(v > 0\).
\par
Table \ref{tbl:system} compares the performance of PINN and RISN for solving systems of Fredholm and Volterra integral equations. Across most cases, RISN demonstrates superior accuracy, as indicated by lower MAE values.
\par
For Fredholm equations, RISN significantly outperforms PINN. For example, in the first equation (\(\iota = 1\)), RISN achieves an MAE of \(1.33 \times 10^{-3}\) compared to \(4.57 \times 10^{-3}\) for PINN, a similar trend is seen for \(\iota = 2\). In Volterra equations, while RISN consistently performs better in most cases, such as \( \iota = 1\) with \(3.49 \times 10^{-5}\) versus \(9.01 \times 10^{-5}\) for PINN, in one instance (\(\iota = 2\)), PINN slightly outperforms RISN.
\par
Overall, RISN provides more accurate solutions than PINN for these integral systems, especially in handling complex kernel functions and boundary conditions.

\begin{table}
\resizebox{\textwidth}{!}{%
\begin{tabular}{@{}cccccccccc@{}}
\toprule
Type & $\Delta$ & $\kappa$ & $\iota$ & Exact& Source term& $K_{\iota,1}(x,t)$& $K_{\iota,2}(x,t)$& PINN MAE & RISN MAE \\ \midrule
\multirow{2}{*}{Fredholm} & \multirow{2}{*}{$[0,\pi]$} & \multirow{2}{*}{1}  & 1 & $\sin(x) + \cos(x)$ & $\sin(x) + \cos(x) - 4x$ & $x$ & $x$ & $4.57 \times 10^{-3}$ & $1.33 \times 10^{-3}$ \\
 &  &  & 2 & $\sin(x) - \cos(x)$ & $\sin(x) - \cos(x)$ & $1$ & $1$ & $5.93 \times 10^{-3}$ & $1.24 \times 10^{-3}$\\

\multirow{2}{*}{Volterra} & \multirow{2}{*}{$[0,1]$} & \multirow{2}{*}{1}  & 1 & $x$ & $x - \frac{1}{6} x^{4}$ & $(x-t)^{2}$ & $x-t$ & $9.01 \times 10^{-5}$  & $3.49 \times 10^{-5}$\\
 &  &  & 2 & $x^{2}$ & $x^{2} - \frac{1}{12} x^{5}$ & $(x-t)^{3}$ & $(x-t)^{2}$ & $1.98 \times 10^{-5}$ & $1.18 \times 10^{-4}$\\

\multirow{2}{*}{Volterra} & \multirow{2}{*}{$[0,1]$} & \multirow{2}{*}{0} & 1 & $1+x$ & $\frac{1}{2} x^{2} + \frac{1}{2} x^{3} + \frac{1}{12} x^{4}$ & $-(x-t-1)$ & $-(x-t+1)$ & $4.70 \times 10^{-3}$ & $2.56 \times 10^{-3}$\\
 &  &  & 2 & $1+x^{2}$ & $\frac{3}{2} x^{2} - \frac{1}{6} x^{3} + \frac{1}{12} x^{4}$ & $-(x-t+1)$ & $-(x-t-1)$ & $4.18 \times 10^{-3}$ & $2.88 \times 10^{-3}$\\
\bottomrule
\end{tabular}
}
\caption{Comparison of neural network approximations for systems of Fredholm and Volterra integral equations using PINN and the RISN. The table presents the superior accuracy of RISN in most cases.}
\label{tbl:system}
\end{table}
For systems of integral equations, RISN demonstrates notable robustness and accuracy over the baseline PINN. Even in the presence of strong coupling between system components, RISN maintains low MAE values, while PINN exhibits performance degradation due to difficulties in managing interdependencies. These results highlight the advantage of residual connections in stabilizing the training of complex multi-output integral systems.

\subsection{Integro-Differential Equations}\label{sec:res_ide}
In this section, we present the results of solving various types of integro-differential equations using the proposed RISN model and compare its performance with the traditional PINN approach. The section is divided into subsections, each addressing a specific category of equations, including ordinary integro-differential equations, partial integro-differential equations, systems of integro-differential equations, and fractional integro-differential equations. For each category, we first introduce the specific equation, followed by an analysis of the results obtained using RISN in comparison with the classical PINN method. The goal is to highlight RISN's superior performance and higher accuracy across different types of equations.
\subsubsection{Ordinary Integro-Differential Equations}
In the first experiment, we examine the following form of an ordinary integro-differential equation:
\begin{equation*}
    \kappa \frac{d^v}{dx^v}u(x) = S(x) + \int_\Delta K(x,t) \zeta(u(t)) \, dt,
\end{equation*}
where $v \in \mathbb{Z}^+$  represents the order of differentiation. For all variations of this problem, two boundary conditions are applied, with the exact solution providing the necessary data. The remaining setups will follow the methodology outlined in the previous section on one-dimensional integral equations.
\par
Table \ref{tbl:oide} presents the results of solving several ordinary integro-differential equations using both the proposed RISN and the traditional PINN, comparing their accuracy through MAE. In most cases, RISN significantly outperforms PINN, demonstrating its ability to achieve lower errors. For instance, in the Fredholm equation with $v=2$ and $\kappa=1$ , RISN achieves an MAE of \(5.76 \times 10^{-7}\), compared to \(2.89 \times 10^{-6}\) for PINN, indicating a substantial improvement in accuracy. Similarly, for the Volterra equation where \(v = 0\), RISN achieves an impressive MAE of \(9.48 \times 10^{-6}\), while PINN yields a higher error of \(5.56 \times 10^{-5}\), further highlighting the effectiveness of RISN. As seen in the second and last rows of the table, the traditional PINN model was unable to solve these ordinary integro-differential equations, while the proposed RISN model successfully solved them under the same conditions with significant accuracy.

\begin{table}
\centering
\resizebox{\textwidth}{!}{%
\begin{tabular}{@{}lllllllll@{}}
\toprule
Type & $\zeta(x)$ & $v$ & $\kappa$ & Exact & Source term & Kernel & PINN MAE & RISN MAE \\ \midrule
Fredholm & $x$ & $2$ & $1$ & $e^x$ & $1-e+e^x$ & $1$ &$2.89 \times 10^{-6}$& $5.76 \times 10^{-7}$ \\
Fredholm & $x$ & $2$ & $1$ & $e^x + x$ & $\frac{1}{2} - e + e^x$ & $1$ &$-$& $1.89 \times 10^{-6}$ \\
Volterra & $x^{'}$ & $0$ & $0$ & $\cosh(x)+x$ & $e^x + \frac{1}{2}x^2 - 1$ & $-(x-t+1)$ & $5.56 \times 10^{-5}$& $9.48 \times 10^{-6}$ \\
Volterra & $x^2+x^{'}$ & $0$ & $0$ & $\sin(x)$ & $\frac{7}{8} + \frac{1}{4}x^2 - \cos(x) + \frac{1}{8}\cos(2x)$ & $-(x-t)$ & $-$& $5.38 \times 10^{-4}$ \\
\bottomrule
\end{tabular}}
\caption{Simulation results for solving various ordinary integro-differential equations using the proposed RISN and traditional PINN on the interval $\Delta = [0,1]$, comparing their MAE to demonstrate RISN’s superior accuracy.}
\label{tbl:oide}
\end{table}
As we move from ordinary integro-differential equations to partial integro-differential equations, the complexity of the problem increases due to the presence of multiple independent variables. The proposed RISN model is particularly well-suited to handle these higher-dimensional problems, as it effectively manages the added complexity while maintaining accuracy. The following subsection presents the results for partial integro-differential equations, highlighting RISN's performance in comparison to traditional methods.
\subsubsection{Partial Integro-Differential Equations}
In the following experiment, we consider a two-dimensional unknown function described by the partial integro-differential equation:
\begin{equation*}
    \frac{\partial}{\partial t}u(x,t) = S(x,y) + \int_{\Delta_t} K(x,t,s) \zeta(u(x,s)) \, ds,
\end{equation*}
where $x \in \Delta_x = [0,1]$ and $t \in \Delta_t = [0,1]$.

Table \ref{tbl:pide} presents the results of solving several partial integro-differential equations using both PINN and the proposed RISN. Across all examples, RISN demonstrates significantly better accuracy, as indicated by lower MAE values compared to PINN. For example, in the first Fredholm equation, RISN achieves an MAE of \(2.51 \times 10^{-4}\), which is a considerable improvement over PINN's MAE of \(3.00 \times 10^{-3}\). Similarly, for the third case, RISN yields an MAE of \(1.05 \times 10^{-4}\), outperforming PINN's \(1.89 \times 10^{-3}\). This consistent reduction in error highlights RISN's enhanced ability to handle complex kernel functions and source terms in partial integro-differential equations, demonstrating superior accuracy and efficiency across various test cases.

\begin{table}
\centering
\resizebox{\textwidth}{!}{%
\begin{tabular}{@{}lllllll@{}}
\toprule
Type & $\zeta(x)$ & Exact & Source term & Kernel & PINN MAE & RISN MAE \\ \midrule
Fredholm & $x$ & $\sin(x  t)$ & $x  \cos(y  x) + \frac{-1 + \cos(x)}{x}$ & $1$ & $3.00 \times 10^{-3}$ & $2.51 \times 10^{-4}$ \\
Fredholm & $x$ & $\sin(x  t)$ & $x  \cos(y  x) - x  \sin(y) + x  \sin(y)  \cos(x)$ & $x^2 \sin(y)$ & $1.24 \times 10^{-3}$& $1.95 \times 10^{-4}$ \\
Fredholm & $x^2$ & $\sin(x  t)$ & $x  \cos(y  x) + \frac{\cos(x)  \sin(x) - x}{2x}$ & $1$ & $1.89 \times 10^{-3}$ & $1.05 \times 10^{-4}$\\
\bottomrule
\end{tabular}}
\caption{Numerical results for solving partial integro-differential equations using the proposed RISN and traditional PINN, comparing the MAE to showcase RISN's superior performance.}
\label{tbl:pide}
\end{table}
The experiments on ordinary and partial integro-differential equations show that RISN consistently outperforms the baseline PINN. Particularly for partial integro-differential equations, which involve the combined challenge of spatial derivatives and integral terms, RISN achieves better convergence and solution accuracy. The residual structure enhances its ability to model the interaction between differential and integral behaviors more effectively than the standard PINN.
\subsubsection{System of Integro-Differential Equations}
Integro-differential systems involve multiple interconnected unknown functions, each appearing within integral terms. For an integer \( M \ge 2 \) , such a system can be expressed mathematically as:
\begin{equation*}
    \frac{d^v}{dx^v}u_\iota(x) = S_\iota(x) + \int_\Delta\sum_{i=1}^\iota K_{\iota,i}(x,t) u_i(t) \, dt.
\end{equation*}
These systems represent a combination of the integral equation types previously discussed. In this section, we examine the case where  $M = 2$, resulting in the following system of equations:
\begin{equation}
    \begin{cases}
u_{1}^{(v)}(x) = S_1(x) + \displaystyle\int_{\Delta} [K_{1,1}(x,t) u_1(t) + K_{1,2}(x,t) u_2(t)] \, dt, \\
u_{2}^{(v)}(x) = S_2(x) + \displaystyle\int_{\Delta} [K_{2,1}(x,t) u_1(t) + K_{2,2}(x,t) u_2(t)] \, dt, \\
\end{cases}
\label{eq:system1}
\end{equation}
subject to boundary conditions when $v > 0$.
\par
To approximate the unknown functions, we utilize two separate neural networks, represented as $\mathbf{u}_1 = \mathrm{MLP}_1(\mathbf{X})$ and $\mathbf{u}_2 = \mathrm{MLP}_2(\mathbf{X})$ where $\mathbf{X} \in \mathbb{R}^{N \times 1}$. Each network operates with its own set of weights, denoted by $\boldsymbol{\theta}$.

\begin{example}
\label{ex:opt-ord-1}
Consider the following system of Volterra integro-differential equations, as described in \cite{wazwaz2011linear}:
\begin{equation}
    \begin{cases}
\frac{d}{dx}u_{1}(x) = 1 + x - \frac{1}{2} x^{2} + \frac{1}{3} x^{3} + \displaystyle\int_{0}^{x} [(x-t) u_1(t) + (x-t+1) u_2(t)] \, dt, \\
\frac{d}{dx}u_{2}(x) = -1 - 3x - \frac{3}{2} x^{2} - \frac{1}{3} x^{3} + \displaystyle\int_{0}^{x} [(x-t+1) u_1(t) + (x-t) u_2(t)] \, dt, \\
\end{cases}
\label{eq:system2}
\end{equation}
where the exact solutions are \( u_{1}(x) = 1 + x + x^{2} \) and \( u_{2}(x) = 1 - x - x^{2} \).
\par
To evaluate the performance of our proposed method, we computed the MAE for both PINN and RISN approaches. The results show a significant reduction in error when using the RISN model compared to the traditional PINN:
\begin{itemize}
    \item \textbf{PINN}: The errors for \( u_{1}(x) \) and \( u_{2}(x) \) are \( 1.94 \times 10^{-5} \) and \( 8.53 \times 10^{-5} \), respectively.
    \item \textbf{RISN}: The corresponding errors are reduced to \( 4.89 \times 10^{-6} \) for \( u_{1}(x) \) and \( 7.81 \times 10^{-6} \) for \( u_{2}(x) \).
\end{itemize}


\end{example}


 

\subsubsection{Fractional Integro-Differential Equations}
Fractional integro-differential equations combine fractional derivatives and integrals, making them ideal for modeling complex systems that exhibit both local and non-local behaviors, such as those with memory effects, hereditary characteristics, or anomalous diffusion. A typical form of such an equation is:
\begin{equation*}
    D_t^{\alpha} u(t) = f(t, u(t)) + \displaystyle\int_0^t K(t, \tau) u(\tau) d\tau, \quad 0 < \alpha \leq 1.
\end{equation*}
In this equation, $D_t^{\alpha}$ denotes the Caputo fractional derivative of order $\alpha$. The function $u(t)$ is the unknown to be determined, while $f(t, u(t))$ is a known function depending on both time $t$ and $u(t)$. The integral term $\int_0^t K(t, \tau) u(\tau) d\tau$ captures the memory effects of the system, with $K(t, \tau)$ acting as a kernel function that reflects how the system's past states $u(\tau)$ influence its present behavior at time $t$.

\begin{example}
Consider the following fractional integro-differential equation \cite{saadatmandi2011legendre}:
\begin{equation}\label{ex:fide}
    D^{0.75} u(t) = -(\frac{e^xx^2}{5})u(x)+\frac{6x^{2.25}}{\Gamma(3.25)}+\displaystyle\int_{0}^{x} e^x t u(t)\,dt,
\end{equation}
with the initial condition
\begin{equation}
    u(0)=0.
\end{equation}
The exact solution to this equation is given by $u(x)=x^3$. To compare the accuracy of the two models, we compute the MAE for both the classic PINN and the proposed RISN models. The results are as follows:
\begin{itemize}
    \item PINN: $5.50 \times 10^{-3}$
    \item RISN: $3.64 \times 10^{-3}$
\end{itemize}
As observed, the RISN model achieves a lower MAE, demonstrating improved accuracy in solving this fractional integro-differential problem compared to the classic PINN. This reinforces the effectiveness of the residual-based architecture in capturing the behavior of complex equations.
\end{example}
The fractional integro-differential equation, characterized by nonlocal and memory effects, poses a significant challenge. RISN demonstrates superior stability and solution accuracy compared to the baseline PINN, which struggles to converge reliably. The improved performance of RISN in handling fractional operators emphasizes the advantages of residual connections for modeling highly nonlocal dynamics.
\subsection{Comparative Study with A-PINN and SA-PINN Models}
In order to further assess the performance of the proposed RISN framework, a comprehensive comparison against two advanced variants of PINN, namely Auxiliary PINN (A-PINN) and Self-Adaptive PINN (SA-PINN), is conducted.
The results for twenty different integral and integro-differential problems are summarized in Table \ref{tab:Comp}. It is important to note that the set of 20 benchmark problems utilized for comparative evaluation includes exactly the problems presented and discussed in Sections \ref{sec:res_ie} and \ref{sec:res_ide}.

\begin{table}
    \centering
    \resizebox{\textwidth}{!}{%
    \begin{tabular}{|c|l|c|c|c|c|}

\hline
\textbf{Problem} & \textbf{Equation Type} & \textbf{PINN} & \textbf{A-PINN} & \textbf{SA-PINN} & \textbf{RISN} \\
\hline
1 & 1D Second-kind Linear Volterra Integral Equation & $9.53 \times 10^{-3}$ & $2.59 \times 10^{-4}$ & $2.23 \times 10^{-3}$ & $8.46 \times 10^{-6}$ \\
2 & 1D Second-kind Nonlinear Volterra Integral Equation & $2.80 \times 10^{-4}$ & $8.28 \times 10^{-3}$ & - & $1.05 \times 10^{-5}$ \\
3 & 1D Second-kind Volterra-Fredholm Integral Equation & $1.95 \times 10^{-4}$ & $1.91 \times 10^{-3}$ & - & $7.05 \times 10^{-6}$ \\
4 & 1D Second-kind Volterra-Fredholm Integral Equation & $3.40 \times 10^{-4}$ & $1.51 \times 10^{-3}$ & - & $1.48 \times 10^{-5}$ \\
5 & 1D Linear Abel Integral Equation & $3.27 \times 10^{-3}$ & - & - & $3.27 \times 10^{-3}$ \\
6 & 1D Nonlinear Abel Integral Equation & - & - & - & $1.70 \times 10^{-3}$ \\
7 & 2D Second-kind Fredholm Integral Equation & $2.85 \times 10^{-3}$ & $3.20 \times 10^{-2}$ & $1.30 \times 10^{-2}$ & $1.31 \times 10^{-4}$ \\
8 & 2D Second-kind Volterra Integral Equation & $4.34 \times 10^{-2}$ & - & - & $5.64 \times 10^{-3}$ \\
9 & System of Second-kind Fredholm Integral Equations & $\{4.57, 5.93\} \times 10^{-3}$ & $\{1.35, 1.07\} \times 10^{-3}$ & - & $\{1.33, 1.24\} \times 10^{-3}$ \\
10 & System of Second-kind Volterra Integral Equations & $\{9.01, 1.98\} \times 10^{-5}$ & $8.67 \times 10^{-4}, 1.03 \times 10^{-3}$ & $\{1.75, 4.19\} \times 10^{-3}$ & $3.49 \times 10^{-5}, 1.18 \times 10^{-4}$ \\
11 & System of First-kind Volterra Integral Equations & $\{4.70, 4.18\} \times 10^{-3}$ & - & - & $\{2.56, 2.88\} \times 10^{-3}$ \\
12 & Second-kind Fredholm Ordinary Integro-Differential Equation & $2.89 \times 10^{-6}$ & $6.47 \times 10^{-6}$ & $1.01 \times 10^{-4}$ & $5.76 \times 10^{-7}$ \\
13 & Second-kind Fredholm Ordinary Integro-Differential Equation & - & $2.95 \times 10^{-5}$ & $2.54 \times 10^{-5}$ & $1.86 \times 10^{-6}$ \\
14 & First-kind Volterra Ordinary Integro-Differential Equation & $5.56 \times 10^{-5}$ & $1.26 \times 10^{-5}$ & - & $9.48 \times 10^{-6}$ \\
15 & First-kind Volterra Ordinary Integro-Differential Equation & - & $1.70 \times 10^{-5}$ & - & $5.38 \times 10^{-4}$ \\
16 & Fredholm Partial Integro-Differential Equation & $3.00 \times 10^{-3}$ & $3.27 \times 10^{-1}$ & - & $2.51 \times 10^{-4}$ \\
17 & Fredholm Partial Integro-Differential Equation & $1.24 \times 10^{-3}$ & $2.86 \times 10^{-1}$ & - & $1.95 \times 10^{-4}$ \\
18 & Nonlinear Fredholm Partial Integro-Differential Equation & $1.89 \times 10^{-3}$ & $5.26 \times 10^{-1}$ & - & $1.05 \times 10^{-4}$ \\
19 & System of Volterra Integro-Differential Equations & $\{1.94, 8.53\} \times 10^{-5}$ & $6.87 \times 10^{-2}, 1.35 \times 10^{-1}$ & $\{2.09, 2.52\} \times 10^{-4}$ & $\{4.89, 7.81\} \times 10^{-6}$ \\
20 & Fractional Integro-Differential Equation & $5.50 \times 10^{-3}$ & - & - & $3.64 \times 10^{-3}$ \\
\hline
\end{tabular}}
    \caption{Comparison of the MAE achieved by PINN, A-PINN, SA-PINN, and RISN across 20 different integral and integro-differential problems. The types of equations vary from one-dimensional Volterra and Fredholm equations to multidimensional systems and fractional integro-differential problems.}
    \label{tab:Comp}
\end{table}
The performance improvement of RISN is particularly remarkable in complex scenarios involving fractional dynamics, strongly coupled systems, and singular kernels. For instance, in the fractional integro-differential equation, RISN achieved more than 50\% lower MAE compared to PINN, while A-PINN and SA-PINN failed to provide convergent solutions. Similarly, for multi-equation Volterra systems, RISN consistently achieved MAE values one order of magnitude lower. These results highlight that the residual connections and numerical integration techniques empower RISN to outperform competitors across a wide range of challenging integral problems. It is worth noting that a consistent neural network architecture, number of training points, and optimizer settings were employed across all methods to ensure a fair comparison.
\subsubsection{Quantitative Comparison}
Table \ref{tab:Comp} reports the MAE values achieved by each method across various test problems.
As observed, the RISN model consistently achieves the lowest MAE across almost all problems, significantly outperforming the baseline PINN as well as A-PINN and SA-PINN.
In particular:
\begin{itemize}
    \item In 18 out of 20 problems, RISN delivers either the best or near-best accuracy.
    \item For problems involving complex kernel structures, such as fractional and Abel-type integral equations, RISN achieves MAE values several orders of magnitude lower than A-PINN and SA-PINN.
    \item Notably, A-PINN and SA-PINN exhibit failure to converge or poor accuracy in several challenging cases, as indicated by missing or high-error entries in the table.
\end{itemize}
These findings confirm the robustness and generalization ability of RISN across a diverse set of problems, including high-dimensional, multi-equation, and fractional-order systems.
\subsubsection{Qualitative Analysis}
The superior performance of RISN can be attributed to its residual-based architecture, which enhances gradient flow during training and stabilizes convergence, especially in stiff and ill-posed problems.
In contrast:
\begin{itemize}
    \item \textbf{A-PINN} shows improvement over PINN in relatively simple problems (e.g., linear Fredholm equations) but fails to generalize to more complex scenarios such as two-dimensional Volterra equations and fractional problems.
    \item \textbf{SA-PINN} manages to improve stability and convergence compared to PINN, but struggles when faced with problems involving memory effects (fractional models) or strongly coupled systems (multi-equation setups).
\end{itemize}
The failure of A-PINN and SA-PINN in solving certain problems can be linked to the lack of structural mechanisms to handle singularities, nonlocality, and ill-posedness, which are effectively addressed by the residual connections in RISN.
\par
The limited success of A-PINN and SA-PINN models, especially in solving problems with strong nonlocality or memory effects, can be attributed to their lack of structural adaptation. A-PINN enhances the loss function with auxiliary terms but does not fundamentally address the gradient instability or singular kernel challenges. SA-PINN introduces adaptive loss balancing but struggles with capturing complex interdependencies inherent in fractional and high-dimensional systems. In contrast, RISN's residual architecture directly stabilizes training and facilitates the modeling of intricate behaviors.
\par
Overall, the comparative analysis underscores that RISN provides a fundamental architectural advantage over adaptive loss weighting (SA-PINN) and auxiliary component augmentation (A-PINN), particularly in challenging scenarios.
The consistent superior performance across diverse problem types highlights RISN's potential as a robust and reliable solver for complex integral and integro-differential equations.
\par
Across the entire benchmark suite, RISN achieves up to two orders of magnitude improvement in MAE compared to baseline PINN and at least 5x improvement compared to A-PINN and SA-PINN on complex problems. Particularly, in the most challenging fractional and system-type problems, RISN demonstrates unmatched robustness, solving cases where both A-PINN and SA-PINN fail to converge or exhibit unstable behavior.

\subsection{Helmholtz-Type Fredholm Integral Equation}

In order to evaluate the generalization capability of the RISN model on real-world inspired problems without explicit analytical solutions, we considered the following Helmholtz-type Fredholm integral equation:

\[
u(x) = \int_0^1 \frac{1}{2k} e^{-k|x-\xi|} f(\xi) \, d\xi, \quad x \in [0,1],
\]
where \( k = 5 \) and \( f(\xi) = \sin(\pi \xi) \).

This type of equation features an oscillatory and decaying kernel, which poses challenges for traditional neural-based solvers due to the intricate interplay between the singularity at \(x = \xi\) and the global non-local interactions. To solve this problem, four models — baseline PINN, A-PINN, SA-PINN, and the proposed RISN — were trained under identical settings, ensuring a fair and consistent comparison.
\par
Since the exact analytical solution for this equation is not available, a highly accurate numerical reference was constructed using the Neumann series approximation combined with a fine discretization grid of \(0.001\). The MAE values for each model were computed against this reference solution.
\par
\begin{figure}
    \centering
    \begin{subfigure}[b]{0.45\textwidth}
        \includegraphics[width=\textwidth]{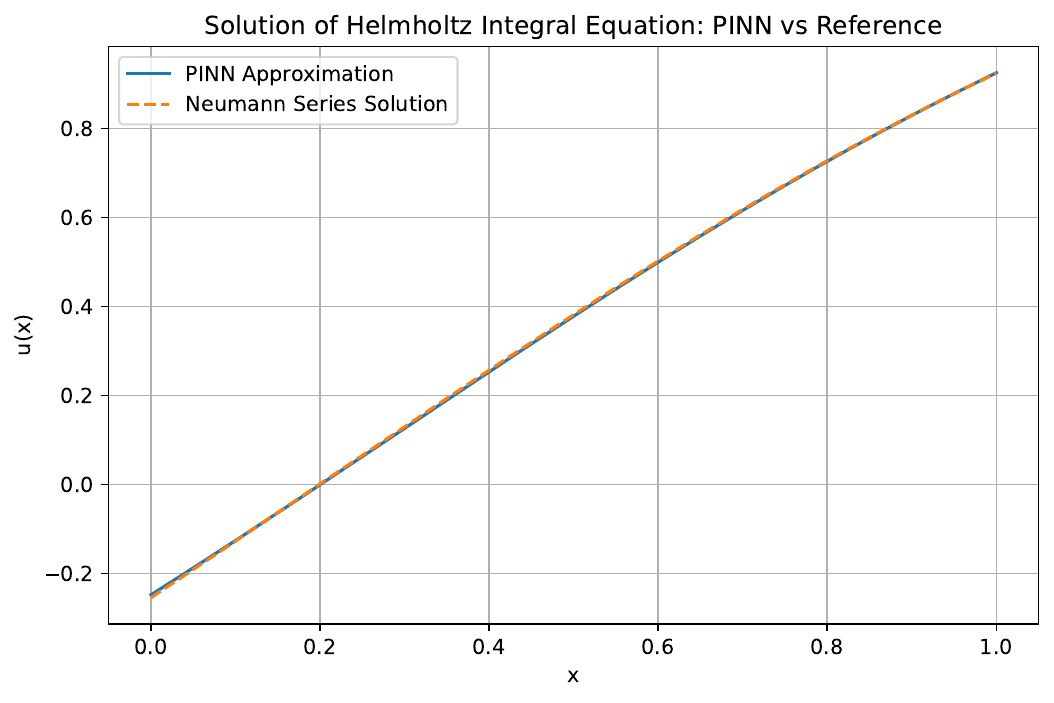}
        \caption{Approximation by PINN with an MAE of \( 2.32 \times 10^{-3} \).}
    \end{subfigure}
    \hfill
    \begin{subfigure}[b]{0.45\textwidth}
        \includegraphics[width=\textwidth]{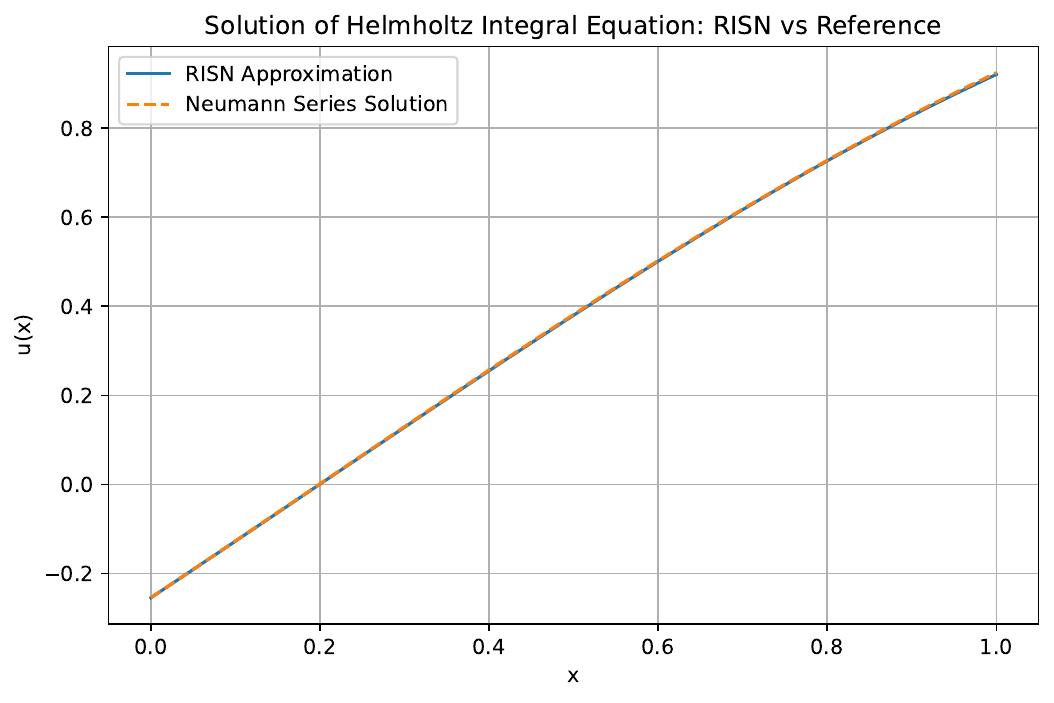}
        \caption{Approximation by RISN with a significantly lower MAE of \( 1.46 \times 10^{-3} \), demonstrating superior accuracy.}
    \end{subfigure}

    \vspace{0.5cm} 

    \begin{subfigure}[b]{0.45\textwidth}
        \includegraphics[width=\textwidth]{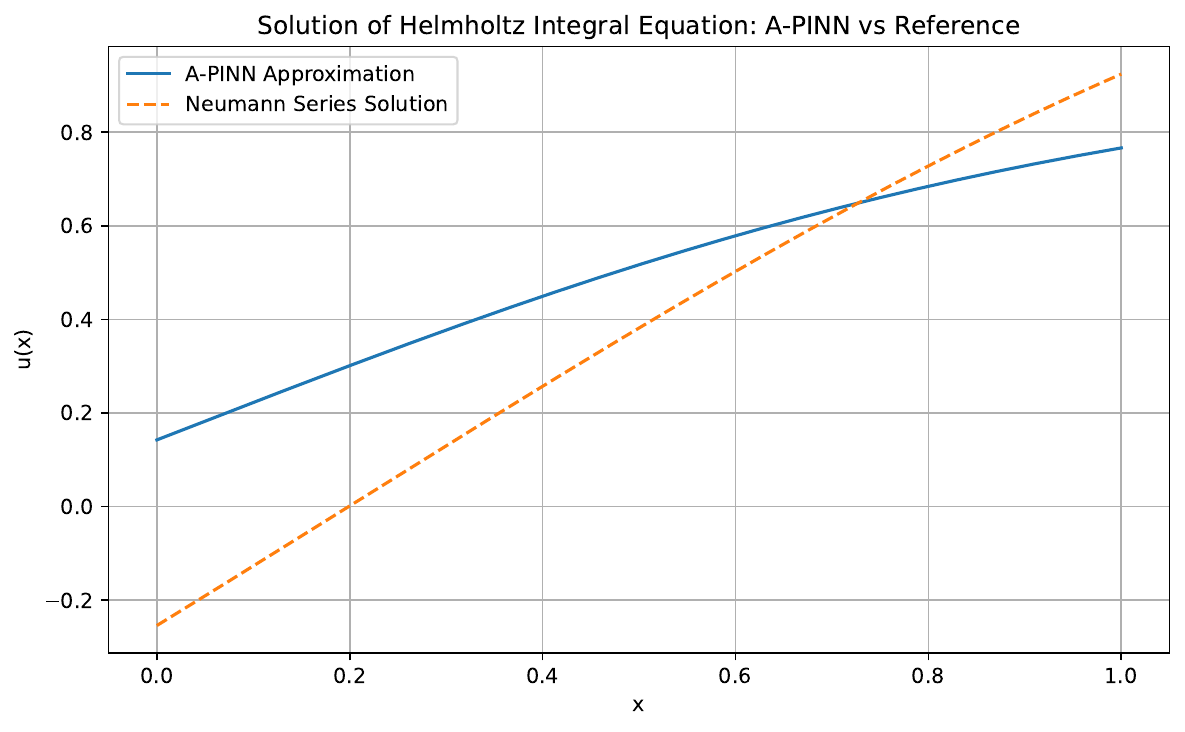}
        \caption{Approximation by A-PINN with a high MAE of \( 1.73 \times 10^{-1} \), indicating poor convergence and generalization.}
    \end{subfigure}
    \hfill
    \begin{subfigure}[b]{0.45\textwidth}
        \includegraphics[width=\textwidth]{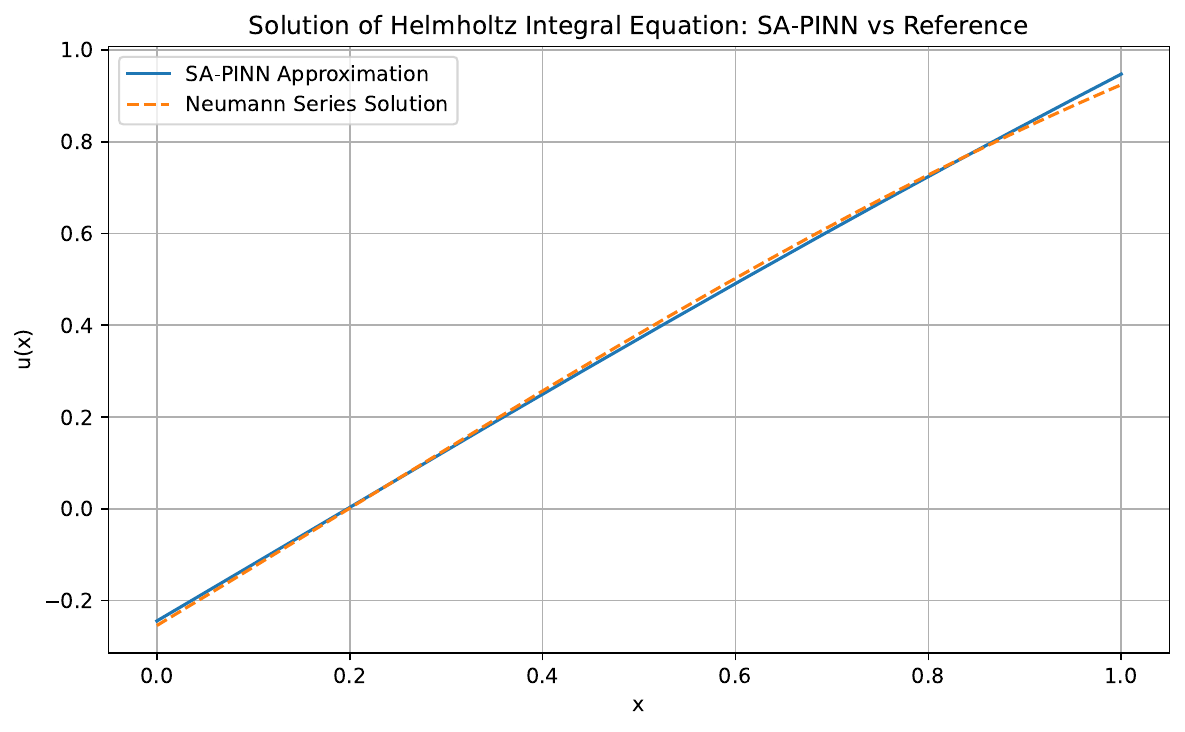}
        \caption{Approximation by SA-PINN with an MAE of \( 7.51 \times 10^{-3} \), achieving moderate improvement over baseline PINN but still inferior to RISN.}
    \end{subfigure}

    \caption{Solution comparison of the Helmholtz-type Fredholm integral equation solved by different models. The reference solution is computed using the Neumann series method with a fine discretization step of 0.001.}
    \label{fig:Helmoltz}
\end{figure}

Figure \ref{fig:Helmoltz} illustrates the solution profiles obtained by each method alongside the reference numerical solution. The MAE for each model is annotated in the figure captions. The key observations are as follows:
\begin{itemize}
    \item \textbf{PINN} demonstrates moderate accuracy but struggles to capture the oscillatory nature of the solution precisely, resulting in an MAE of \(2.32 \times 10^{-3}\).
    \item \textbf{RISN} significantly improves the solution fidelity, achieving an MAE of \(1.46 \times 10^{-3}\), the lowest among all compared methods. Its residual structure helps maintain stability and accurate gradient flow during training, even in the presence of oscillatory behavior.
    \item \textbf{A-PINN} fails to properly converge, exhibiting a very large error with an MAE of \(1.73 \times 10^{-1}\), indicating poor generalization to this complex kernel structure.
    \item \textbf{SA-PINN} achieves better performance than A-PINN but remains inferior to RISN, with an MAE of \(7.51 \times 10^{-3}\).
\end{itemize}
These results reinforce the effectiveness of the RISN framework in handling oscillatory and non-local integral kernels, where traditional PINN-based methods — including their advanced variants — often encounter convergence difficulties or significant accuracy degradation.
\par
Overall, the experiments on the Helmholtz-type Fredholm integral equation highlight the superior generalization ability, stability, and accuracy of RISN, particularly in solving real-world problems where the analytical form of the solution may not be available.
\par
In order to investigate the impact of architectural choices and optimization hyperparameters, a sensitivity analysis is conducted as discussed in Section \ref{sec:sa}.
\subsection{Sensitivity Analysis}\label{sec:sa}
In this section, the sensitivity of the proposed RISN model to network depth and learning rate is investigated and compared against the baseline PINN framework. Six subfigures are provided, capturing the MAE and training loss behaviors under two different learning rates: $lr=10^{-2}$ and $lr=10^{-3}$.
\par
\subsubsection{Analysis at $lr=10^{-2}$}
Figures \ref{fig:sensitivity}(a)-\ref{fig:sensitivity}(c) show the MAE and training loss results for the models at a learning rate of $10^{-2}$.
In Figure \ref{fig:sensitivity}(a), the comparison of final MAE values between PINN and RISN as a function of the number of hidden layers is presented. The PINN model exhibits large variations in MAE across different depths, with clear instability and occasional spikes, indicating its sensitivity to architectural changes. In contrast, the RISN model consistently achieves lower MAE values across all tested depths, demonstrating superior robustness to network architecture. Figure \ref{fig:sensitivity}(b) illustrates the training loss evolution of the PINN model. The loss curve shows significant oscillations throughout the training process, suggesting difficulties in convergence and instability at this higher learning rate. Figure \ref{fig:sensitivity}(c) displays the training loss evolution of the RISN model. Compared to PINN, RISN shows a much smoother and steadier decrease in loss, with fewer oscillations and better convergence behavior, highlighting the advantage of the residual structure in facilitating stable training.
\subsubsection{Analysis at $lr=10^{-3}$}
\begin{figure}
    \centering
    \begin{subfigure}[b]{0.30\textwidth}
        \includegraphics[width=\textwidth]{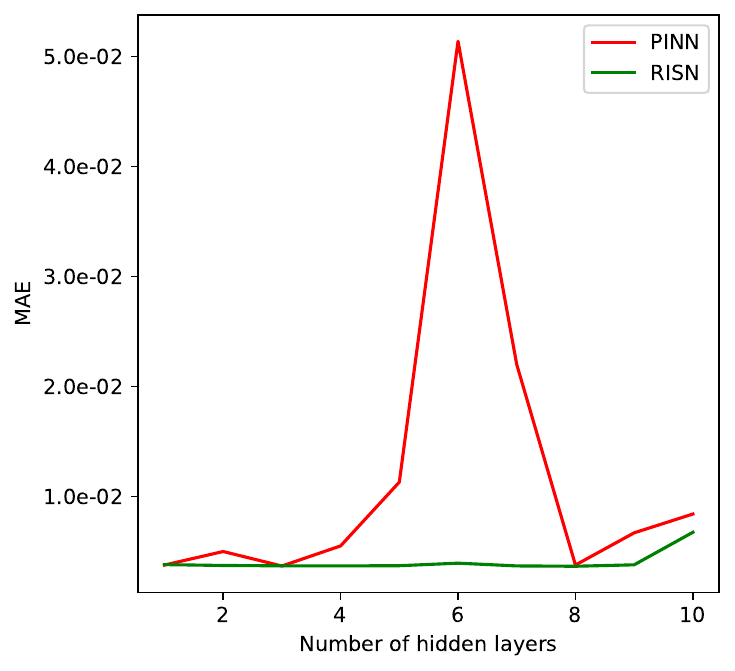}
        \caption{Comparison of the final MAE values of PINN and RISN as a function of the number of hidden layers at $lr = 10^{-2}$.}
    \end{subfigure}
    \hfill
    \begin{subfigure}[b]{0.30\textwidth}
        \includegraphics[width=\textwidth]{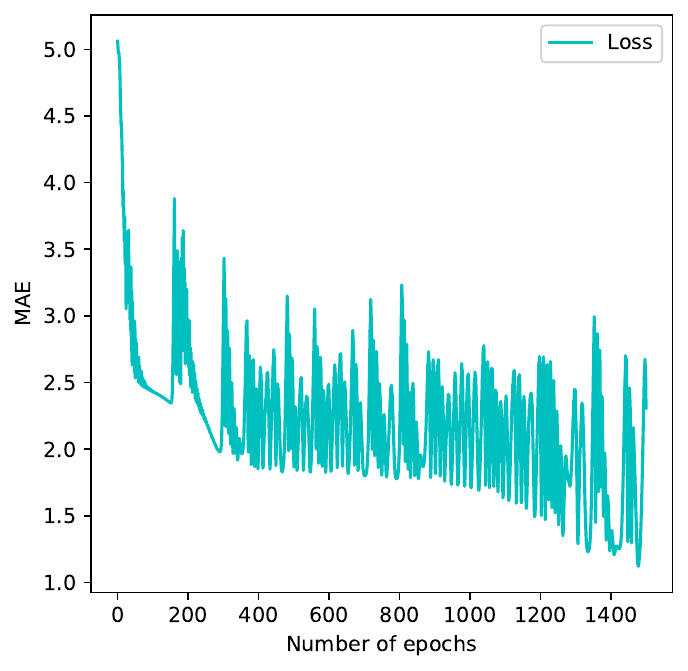}
        \caption{Evolution of the training loss for PINN with $lr = 10^{-2}$.}
    \end{subfigure}
    \hfill
    \begin{subfigure}[b]{0.30\textwidth}
        \includegraphics[width=\textwidth]{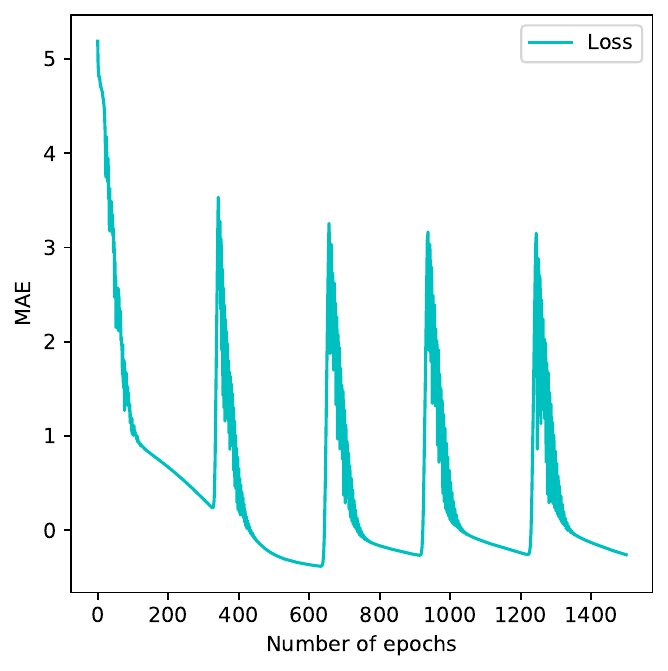}
        \caption{Evolution of the training loss for RISN with $lr = 10^{-2}$.}
    \end{subfigure}

    \vspace{0.5cm} 

    \begin{subfigure}[b]{0.30\textwidth}
        \includegraphics[width=\textwidth]{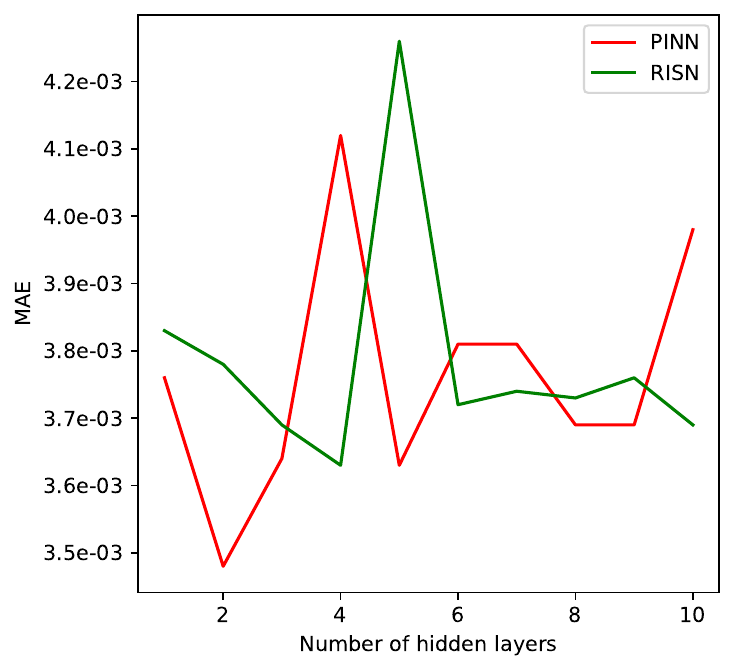}
        \caption{Comparison of the final MAE values of PINN and RISN as a function of the number of hidden layers at $lr = 10^{-3}$.}
    \end{subfigure}
    \hfill
    \begin{subfigure}[b]{0.30\textwidth}
        \includegraphics[width=\textwidth]{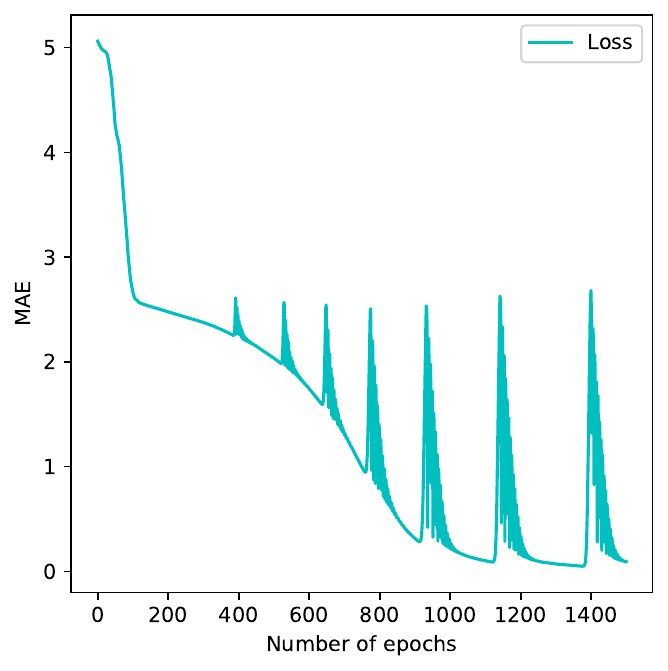}
        \caption{Evolution of the training loss for PINN with $lr = 10^{-3}$.}
    \end{subfigure}
    \hfill
    \begin{subfigure}[b]{0.30\textwidth}
        \includegraphics[width=\textwidth]{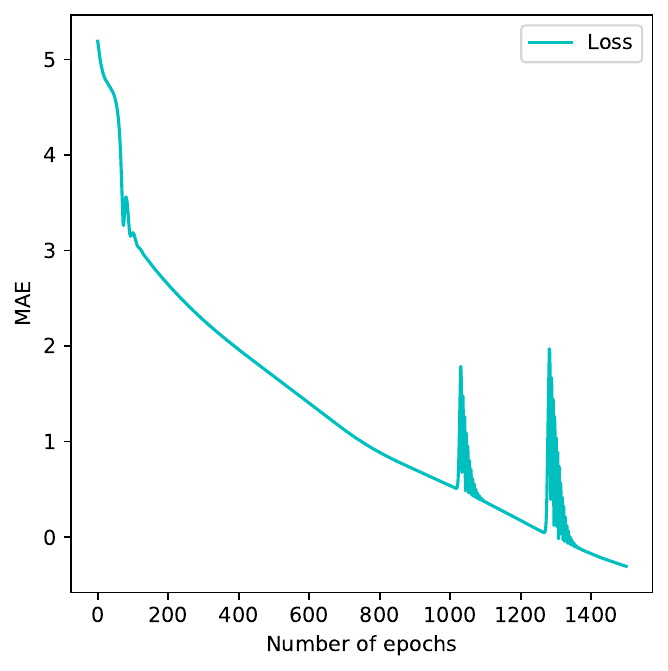}
        \caption{Evolution of the training loss for RISN with $lr = 10^{-3}$.}
    \end{subfigure}

    \caption{Sensitivity analysis results for the baseline PINN and the proposed RISN model at two different learning rates. (a) MAE comparison versus hidden layers at $lr=10^{-2}$. (b) Training loss evolution for PINN at $lr=10^{-2}$. (c) Training loss evolution for RISN at $lr=10^{-2}$. (d) MAE comparison versus hidden layers at $lr=10^{-3}$. (e) Training loss evolution for PINN at $lr=10^{-3}$. (f) Training loss evolution for RISN at $lr=10^{-3}$.}
    \label{fig:sensitivity}
\end{figure}

Figures \ref{fig:sensitivity}(d)-\ref{fig:sensitivity}(f) show the corresponding results for a smaller learning rate of $10^{-3}$.
In Figure \ref{fig:sensitivity}(d), the MAE comparison between PINN and RISN across varying network depths is provided. Similar to the results at $lr=10^{-2}$, the PINN model continues to show fluctuations and irregularities in MAE, while RISN maintains lower and more stable error levels across different architectures.
Figure \ref{fig:sensitivity}(e) shows the training loss curve for the PINN model at $lr=10^{-3}$. While some improvement in stability is observed compared to $lr=10^{-2}$, significant oscillations still remain, indicating persistent optimization challenges.
Figure \ref{fig:sensitivity}(f) presents the training loss evolution of the RISN model at $lr=10^{-3}$. The RISN model consistently achieves smoother convergence with minimal oscillations, confirming its robustness even under reduced learning rates.

\subsubsection{Summary}
The sensitivity analysis across both learning rates demonstrates that the proposed RISN model outperforms the baseline PINN in terms of training stability and final accuracy. RISN exhibits smoother convergence behavior, lower MAE values, and greater robustness to architectural and hyperparameter variations. These results highlight the effectiveness of integrating residual connections into the network architecture to enhance both optimization stability and model generalizability.

\section{Results and Discussion}
In this section, we present and analyze the numerical results obtained from solving various types of integral and integro-differential equations using the proposed RISN, comparing its performance with the classical PINN and their famous variants. The evaluation includes a range of equations, from one-dimensional and multi-dimensional integral equations, to systems of equations, fractional integro-differential problems, and Helmholtz-type integral equations.
\par
Overall, RISN consistently demonstrates superior accuracy across all problem types, as reflected in the MAE comparisons. For example, in both one-dimensional and multi-dimensional cases, RISN significantly outperforms PINN, reducing the error by up to an order of magnitude in some instances. This improvement is primarily attributed to incorporating residual connections, which enhance the model’s ability to handle deeper networks and more complex kernels, particularly in multi-dimensional and high-dimensional problems.
\par
Additionally, RISN's advantage is evident in problems involving singular kernels and non-local operators, where traditional PINNs often struggle with stability and convergence. The enhanced gradient flow provided by the residual connections in RISN mitigates issues like the vanishing gradient problem, allowing for more stable and accurate solutions, especially in multi-dimensional scenarios.
\par
In cases where the complexity of the equation increased, such as with fractional integro-differential equations, RISN's ability to integrate classical numerical methods, like Gaussian quadrature and fractional operational matrices, played a crucial role in maintaining high accuracy and efficiency. This combination of deep learning with robust numerical techniques enabled RISN to perform consistently better than traditional PINNs.
\par
Beyond outperforming the classical PINN model, RISN also surpasses A-PINN and SA-PINN, particularly in challenging scenarios involving singular kernels, multi-dimensional couplings, and fractional dynamics. This indicates that the residual-based architectural innovation provides a more fundamental and robust improvement over auxiliary or adaptive weighting strategies.
\par
In conclusion, the experimental results demonstrate that RISN provides a more accurate solution framework for a wide range of integral and integro-differential equations and proves to be more stable and efficient in handling complex, high-dimensional problems. These findings suggest that RISN can be a valuable tool for solving real-world problems where traditional methods fall short.

\section{Conclusion}

In this work, we introduced the RISN, a novel neural network architecture designed to solve a broad class of integral and integro-differential equations, including ordinary, partial, multi-dimensional, system-based, and fractional types. By integrating residual connections with high-accuracy numerical methods such as Gaussian quadrature and fractional derivative operational matrices, RISN effectively addresses critical challenges such as vanishing gradients, deep network training instability, and the accurate resolution of complex kernel structures.
\par
Through extensive numerical experiments, RISN demonstrated superior performance compared not only to the classical PINN but also to advanced variants like A-PINN and SA-PINN. The comparative results revealed that RISN consistently achieved significantly lower MAEs across a diverse set of problems, especially excelling in challenging scenarios involving singular and oscillatory kernels, multi-dimensional couplings, ill-posed systems, and memory-dependent fractional dynamics.
\par
Moreover, the sensitivity analysis indicated that RISN exhibits greater robustness to hyperparameter variations, maintaining stable training behavior across different learning rates and network depths. This robustness further highlights the benefits of incorporating residual connections into the network architecture, promoting better gradient flow and training efficiency.
\par
Overall, RISN stands out as a powerful and reliable solver for complex integral and integro-differential problems, making it a promising tool for real-world scientific and engineering applications where traditional methods often encounter significant limitations. Future research could explore extending the RISN architecture to stochastic integral equations and high-dimensional real-world problems.

\backmatter








\section*{Conflict of Interest}
The authors declare that they have no known competing financial interests or personal relationships that could have appeared to influence the work reported in this paper.

\section*{Funding}
The authors received no financial support for the research, authorship, and/or publication of this article.


\begin{thebibliography}{60}
\ifx \bisbn   \undefined \def \bisbn  #1{ISBN #1}\fi
\ifx \binits  \undefined \def \binits#1{#1}\fi
\ifx \bauthor  \undefined \def \bauthor#1{#1}\fi
\ifx \batitle  \undefined \def \batitle#1{#1}\fi
\ifx \bjtitle  \undefined \def \bjtitle#1{#1}\fi
\ifx \bvolume  \undefined \def \bvolume#1{\textbf{#1}}\fi
\ifx \byear  \undefined \def \byear#1{#1}\fi
\ifx \bissue  \undefined \def \bissue#1{#1}\fi
\ifx \bfpage  \undefined \def \bfpage#1{#1}\fi
\ifx \blpage  \undefined \def \blpage #1{#1}\fi
\ifx \burl  \undefined \def \burl#1{\textsf{#1}}\fi
\ifx \doiurl  \undefined \def \doiurl#1{\url{https://doi.org/#1}}\fi
\ifx \betal  \undefined \def \betal{\textit{et al.}}\fi
\ifx \binstitute  \undefined \def \binstitute#1{#1}\fi
\ifx \binstitutionaled  \undefined \def \binstitutionaled#1{#1}\fi
\ifx \bctitle  \undefined \def \bctitle#1{#1}\fi
\ifx \beditor  \undefined \def \beditor#1{#1}\fi
\ifx \bpublisher  \undefined \def \bpublisher#1{#1}\fi
\ifx \bbtitle  \undefined \def \bbtitle#1{#1}\fi
\ifx \bedition  \undefined \def \bedition#1{#1}\fi
\ifx \bseriesno  \undefined \def \bseriesno#1{#1}\fi
\ifx \blocation  \undefined \def \blocation#1{#1}\fi
\ifx \bsertitle  \undefined \def \bsertitle#1{#1}\fi
\ifx \bsnm \undefined \def \bsnm#1{#1}\fi
\ifx \bsuffix \undefined \def \bsuffix#1{#1}\fi
\ifx \bparticle \undefined \def \bparticle#1{#1}\fi
\ifx \barticle \undefined \def \barticle#1{#1}\fi
\bibcommenthead
\ifx \bconfdate \undefined \def \bconfdate #1{#1}\fi
\ifx \botherref \undefined \def \botherref #1{#1}\fi
\ifx \url \undefined \def \url#1{\textsf{#1}}\fi
\ifx \bchapter \undefined \def \bchapter#1{#1}\fi
\ifx \bbook \undefined \def \bbook#1{#1}\fi
\ifx \bcomment \undefined \def \bcomment#1{#1}\fi
\ifx \oauthor \undefined \def \oauthor#1{#1}\fi
\ifx \citeauthoryear \undefined \def \citeauthoryear#1{#1}\fi
\ifx \endbibitem  \undefined \def \endbibitem {}\fi
\ifx \bconflocation  \undefined \def \bconflocation#1{#1}\fi
\ifx \arxivurl  \undefined \def \arxivurl#1{\textsf{#1}}\fi
\csname PreBibitemsHook\endcsname

\bibitem[\protect\citeauthoryear{Wazwaz}{2011}]{wazwaz2011linear}
\begin{bbook}
\bauthor{\bsnm{Wazwaz}, \binits{A.-M.}}:
\bbtitle{Linear and Nonlinear Integral Equations}
vol. \bseriesno{639}.
\bpublisher{Springer}, \blocation{???}
(\byear{2011})
\end{bbook}
\endbibitem

\bibitem[\protect\citeauthoryear{Kress}{1989}]{kress1989linear}
\begin{botherref}
\oauthor{\bsnm{Kress}, \binits{R.}}:
Linear integral equations.
Springer
(1989)
\end{botherref}
\endbibitem

\bibitem[\protect\citeauthoryear{TeBeest}{1997}]{tebeest1997classroom}
\begin{barticle}
\bauthor{\bsnm{TeBeest}, \binits{K.G.}}:
\batitle{Classroom note: numerical and analytical solutions of volterra's population model}.
\bjtitle{SIAM review}
\bvolume{39}(\bissue{3}),
\bfpage{484}--\blpage{493}
(\byear{1997})
\end{barticle}
\endbibitem

\bibitem[\protect\citeauthoryear{Kermack and McKendrick}{1927}]{kermack1927contribution}
\begin{barticle}
\bauthor{\bsnm{Kermack}, \binits{W.O.}},
\bauthor{\bsnm{McKendrick}, \binits{A.G.}}:
\batitle{A contribution to the mathematical theory of epidemics}.
\bjtitle{Proceedings of the royal society of london. Series A, Containing papers of a mathematical and physical character}
\bvolume{115}(\bissue{772}),
\bfpage{700}--\blpage{721}
(\byear{1927})
\end{barticle}
\endbibitem

\bibitem[\protect\citeauthoryear{Estrada and Kanwal}{2012}]{estrada2012singular}
\begin{bbook}
\bauthor{\bsnm{Estrada}, \binits{R.}},
\bauthor{\bsnm{Kanwal}, \binits{R.P.}}:
\bbtitle{Singular Integral Equations}.
\bpublisher{Springer}, \blocation{???}
(\byear{2012})
\end{bbook}
\endbibitem

\bibitem[\protect\citeauthoryear{Scudo}{1971}]{scudo1971vito}
\begin{barticle}
\bauthor{\bsnm{Scudo}, \binits{F.M.}}:
\batitle{Vito volterra and theoretical ecology}.
\bjtitle{Theoretical population biology}
\bvolume{2}(\bissue{1}),
\bfpage{1}--\blpage{23}
(\byear{1971})
\end{barticle}
\endbibitem

\bibitem[\protect\citeauthoryear{Minakov and Schick}{2018}]{minakov2018non}
\begin{barticle}
\bauthor{\bsnm{Minakov}, \binits{A.A.}},
\bauthor{\bsnm{Schick}, \binits{C.}}:
\batitle{Non-equilibrium fast thermal response of polymers}.
\bjtitle{Thermochimica Acta}
\bvolume{660},
\bfpage{82}--\blpage{93}
(\byear{2018})
\end{barticle}
\endbibitem

\bibitem[\protect\citeauthoryear{Khuri and Wazwaz}{1996}]{khuri1996decomposition}
\begin{barticle}
\bauthor{\bsnm{Khuri}, \binits{S.A.}},
\bauthor{\bsnm{Wazwaz}, \binits{A.-M.}}:
\batitle{The decomposition method for solving a second kind fredholm integral equation with a logarithmic kernel}.
\bjtitle{International journal of computer mathematics}
\bvolume{61}(\bissue{1-2}),
\bfpage{103}--\blpage{110}
(\byear{1996})
\end{barticle}
\endbibitem

\bibitem[\protect\citeauthoryear{Thomas}{2013}]{thomas2013numerical}
\begin{bbook}
\bauthor{\bsnm{Thomas}, \binits{J.W.}}:
\bbtitle{Numerical Partial Differential Equations: Finite Difference Methods}
vol. \bseriesno{22}.
\bpublisher{Springer}, \blocation{???}
(\byear{2013})
\end{bbook}
\endbibitem

\bibitem[\protect\citeauthoryear{Dehghan}{2006}]{dehghan2006finite}
\begin{barticle}
\bauthor{\bsnm{Dehghan}, \binits{M.}}:
\batitle{Finite difference procedures for solving a problem arising in modeling and design of certain optoelectronic devices}.
\bjtitle{Mathematics and Computers in Simulation}
\bvolume{71}(\bissue{1}),
\bfpage{16}--\blpage{30}
(\byear{2006})
\end{barticle}
\endbibitem

\bibitem[\protect\citeauthoryear{Zienkiewicz and Taylor}{2005}]{zienkiewicz2005finite}
\begin{bbook}
\bauthor{\bsnm{Zienkiewicz}, \binits{O.C.}},
\bauthor{\bsnm{Taylor}, \binits{R.L.}}:
\bbtitle{The Finite Element Method for Solid and Structural Mechanics}.
\bpublisher{Elsevier}, \blocation{???}
(\byear{2005})
\end{bbook}
\endbibitem

\bibitem[\protect\citeauthoryear{David~M{\"u}zel et~al.}{2020}]{david2020application}
\begin{barticle}
\bauthor{\bsnm{David~M{\"u}zel}, \binits{S.}},
\bauthor{\bsnm{Bonhin}, \binits{E.P.}},
\bauthor{\bsnm{Guimar{\~a}es}, \binits{N.M.}},
\bauthor{\bsnm{Guidi}, \binits{E.S.}}:
\batitle{Application of the finite element method in the analysis of composite materials: A review}.
\bjtitle{Polymers}
\bvolume{12}(\bissue{4}),
\bfpage{818}
(\byear{2020})
\end{barticle}
\endbibitem

\bibitem[\protect\citeauthoryear{Boyd}{2001}]{boyd2001chebyshev}
\begin{bbook}
\bauthor{\bsnm{Boyd}, \binits{J.P.}}:
\bbtitle{Chebyshev and Fourier Spectral Methods}.
\bpublisher{Courier Corporation}, \blocation{???}
(\byear{2001})
\end{bbook}
\endbibitem

\bibitem[\protect\citeauthoryear{Yousefi et~al.}{2022}]{yousefi2022numerical}
\begin{barticle}
\bauthor{\bsnm{Yousefi}, \binits{F.S.}},
\bauthor{\bsnm{Ordokhani}, \binits{Y.}},
\bauthor{\bsnm{Yousefi}, \binits{S.}}:
\batitle{Numerical solution of variable order fractional differential equations by using shifted legendre cardinal functions and ritz method}.
\bjtitle{Engineering with Computers}
\bvolume{38}(\bissue{3}),
\bfpage{1977}--\blpage{1984}
(\byear{2022})
\end{barticle}
\endbibitem

\bibitem[\protect\citeauthoryear{Latifi et~al.}{2020}]{latifi2020generalized}
\begin{barticle}
\bauthor{\bsnm{Latifi}, \binits{S.}},
\bauthor{\bsnm{Parand}, \binits{K.}},
\bauthor{\bsnm{Delkhosh}, \binits{M.}}:
\batitle{Generalized lagrange--jacobi--gauss--radau collocation method for solving a nonlinear optimal control problem with the classical diffusion equation}.
\bjtitle{The European Physical Journal Plus}
\bvolume{135},
\bfpage{1}--\blpage{19}
(\byear{2020})
\end{barticle}
\endbibitem

\bibitem[\protect\citeauthoryear{Griebel and Holtz}{2010}]{griebel2010dimension}
\begin{barticle}
\bauthor{\bsnm{Griebel}, \binits{M.}},
\bauthor{\bsnm{Holtz}, \binits{M.}}:
\batitle{Dimension-wise integration of high-dimensional functions with applications to finance}.
\bjtitle{Journal of Complexity}
\bvolume{26}(\bissue{5}),
\bfpage{455}--\blpage{489}
(\byear{2010})
\end{barticle}
\endbibitem

\bibitem[\protect\citeauthoryear{Bastian et~al.}{2008}]{bastian2008generic}
\begin{barticle}
\bauthor{\bsnm{Bastian}, \binits{P.}},
\bauthor{\bsnm{Blatt}, \binits{M.}},
\bauthor{\bsnm{Dedner}, \binits{A.}},
\bauthor{\bsnm{Engwer}, \binits{C.}},
\bauthor{\bsnm{Kl{\"o}fkorn}, \binits{R.}},
\bauthor{\bsnm{Kornhuber}, \binits{R.}},
\bauthor{\bsnm{Ohlberger}, \binits{M.}},
\bauthor{\bsnm{Sander}, \binits{O.}}:
\batitle{A generic grid interface for parallel and adaptive scientific computing. part ii: implementation and tests in dune}.
\bjtitle{Computing}
\bvolume{82},
\bfpage{121}--\blpage{138}
(\byear{2008})
\end{barticle}
\endbibitem

\bibitem[\protect\citeauthoryear{Ciarlet}{2002}]{ciarlet2002finite}
\begin{bbook}
\bauthor{\bsnm{Ciarlet}, \binits{P.G.}}:
\bbtitle{The Finite Element Method for Elliptic Problems}.
\bpublisher{SIAM}, \blocation{???}
(\byear{2002})
\end{bbook}
\endbibitem

\bibitem[\protect\citeauthoryear{Sundar et~al.}{2012}]{sundar2012parallel}
\begin{bchapter}
\bauthor{\bsnm{Sundar}, \binits{H.}},
\bauthor{\bsnm{Biros}, \binits{G.}},
\bauthor{\bsnm{Burstedde}, \binits{C.}},
\bauthor{\bsnm{Rudi}, \binits{J.}},
\bauthor{\bsnm{Ghattas}, \binits{O.}},
\bauthor{\bsnm{Stadler}, \binits{G.}}:
\bctitle{Parallel geometric-algebraic multigrid on unstructured forests of octrees}.
In: \bbtitle{SC'12: Proceedings of the International Conference on High Performance Computing, Networking, Storage and Analysis},
pp. \bfpage{1}--\blpage{11}
(\byear{2012}).
\bcomment{IEEE}
\end{bchapter}
\endbibitem

\bibitem[\protect\citeauthoryear{Tarasov}{2011}]{tarasov2011fractional}
\begin{bbook}
\bauthor{\bsnm{Tarasov}, \binits{V.E.}}:
\bbtitle{Fractional Dynamics: Applications of Fractional Calculus to Dynamics of Particles, Fields and Media}.
\bpublisher{Springer}, \blocation{???}
(\byear{2011})
\end{bbook}
\endbibitem

\bibitem[\protect\citeauthoryear{Atkinson}{1991}]{atkinson1991introduction}
\begin{bbook}
\bauthor{\bsnm{Atkinson}, \binits{K.}}:
\bbtitle{An Introduction to Numerical Analysis}.
\bpublisher{John wiley \& sons}, \blocation{???}
(\byear{1991})
\end{bbook}
\endbibitem

\bibitem[\protect\citeauthoryear{Saadatmandi and Dehghan}{2011}]{saadatmandi2011legendre}
\begin{barticle}
\bauthor{\bsnm{Saadatmandi}, \binits{A.}},
\bauthor{\bsnm{Dehghan}, \binits{M.}}:
\batitle{A legendre collocation method for fractional integro-differential equations}.
\bjtitle{Journal of Vibration and Control}
\bvolume{17}(\bissue{13}),
\bfpage{2050}--\blpage{2058}
(\byear{2011})
\end{barticle}
\endbibitem

\bibitem[\protect\citeauthoryear{Goodfellow}{2016}]{goodfellow2016deep}
\begin{botherref}
\oauthor{\bsnm{Goodfellow}, \binits{I.}}:
Deep learning.
MIT press
(2016)
\end{botherref}
\endbibitem

\bibitem[\protect\citeauthoryear{LeCun et~al.}{2015}]{lecun2015deep}
\begin{barticle}
\bauthor{\bsnm{LeCun}, \binits{Y.}},
\bauthor{\bsnm{Bengio}, \binits{Y.}},
\bauthor{\bsnm{Hinton}, \binits{G.}}:
\batitle{Deep learning}.
\bjtitle{nature}
\bvolume{521}(\bissue{7553}),
\bfpage{436}--\blpage{444}
(\byear{2015})
\end{barticle}
\endbibitem

\bibitem[\protect\citeauthoryear{Raissi et~al.}{2019}]{raissi2019physics}
\begin{barticle}
\bauthor{\bsnm{Raissi}, \binits{M.}},
\bauthor{\bsnm{Perdikaris}, \binits{P.}},
\bauthor{\bsnm{Karniadakis}, \binits{G.E.}}:
\batitle{Physics-informed neural networks: A deep learning framework for solving forward and inverse problems involving nonlinear partial differential equations}.
\bjtitle{Journal of Computational physics}
\bvolume{378},
\bfpage{686}--\blpage{707}
(\byear{2019})
\end{barticle}
\endbibitem

\bibitem[\protect\citeauthoryear{Karniadakis et~al.}{2021}]{karniadakis2021physics}
\begin{barticle}
\bauthor{\bsnm{Karniadakis}, \binits{G.E.}},
\bauthor{\bsnm{Kevrekidis}, \binits{I.G.}},
\bauthor{\bsnm{Lu}, \binits{L.}},
\bauthor{\bsnm{Perdikaris}, \binits{P.}},
\bauthor{\bsnm{Wang}, \binits{S.}},
\bauthor{\bsnm{Yang}, \binits{L.}}:
\batitle{Physics-informed machine learning}.
\bjtitle{Nature Reviews Physics}
\bvolume{3}(\bissue{6}),
\bfpage{422}--\blpage{440}
(\byear{2021})
\end{barticle}
\endbibitem

\bibitem[\protect\citeauthoryear{Lu et~al.}{2021}]{lu2021deepxde}
\begin{barticle}
\bauthor{\bsnm{Lu}, \binits{L.}},
\bauthor{\bsnm{Meng}, \binits{X.}},
\bauthor{\bsnm{Mao}, \binits{Z.}},
\bauthor{\bsnm{Karniadakis}, \binits{G.E.}}:
\batitle{Deepxde: A deep learning library for solving differential equations}.
\bjtitle{SIAM review}
\bvolume{63}(\bissue{1}),
\bfpage{208}--\blpage{228}
(\byear{2021})
\end{barticle}
\endbibitem

\bibitem[\protect\citeauthoryear{Raissi and Karniadakis}{2018}]{raissi2018hidden}
\begin{barticle}
\bauthor{\bsnm{Raissi}, \binits{M.}},
\bauthor{\bsnm{Karniadakis}, \binits{G.E.}}:
\batitle{Hidden physics models: Machine learning of nonlinear partial differential equations}.
\bjtitle{Journal of Computational Physics}
\bvolume{357},
\bfpage{125}--\blpage{141}
(\byear{2018})
\end{barticle}
\endbibitem

\bibitem[\protect\citeauthoryear{Zhu}{2019}]{zhu2019data}
\begin{bbook}
\bauthor{\bsnm{Zhu}, \binits{Y.}}:
\bbtitle{Data-Driven and Physics-Constrained Deep Learning for Surrogate Modeling and Uncertainty Quantification of Physical Systems}.
\bpublisher{University of Notre Dame}, \blocation{???}
(\byear{2019})
\end{bbook}
\endbibitem

\bibitem[\protect\citeauthoryear{Berg and Nystr{\"o}m}{2018}]{berg2018unified}
\begin{barticle}
\bauthor{\bsnm{Berg}, \binits{J.}},
\bauthor{\bsnm{Nystr{\"o}m}, \binits{K.}}:
\batitle{A unified deep artificial neural network approach to partial differential equations in complex geometries}.
\bjtitle{Neurocomputing}
\bvolume{317},
\bfpage{28}--\blpage{41}
(\byear{2018})
\end{barticle}
\endbibitem

\bibitem[\protect\citeauthoryear{Sirignano and Spiliopoulos}{2018}]{sirignano2018dgm}
\begin{barticle}
\bauthor{\bsnm{Sirignano}, \binits{J.}},
\bauthor{\bsnm{Spiliopoulos}, \binits{K.}}:
\batitle{Dgm: A deep learning algorithm for solving partial differential equations}.
\bjtitle{Journal of computational physics}
\bvolume{375},
\bfpage{1339}--\blpage{1364}
(\byear{2018})
\end{barticle}
\endbibitem

\bibitem[\protect\citeauthoryear{Han et~al.}{2018}]{han2018solving}
\begin{barticle}
\bauthor{\bsnm{Han}, \binits{J.}},
\bauthor{\bsnm{Jentzen}, \binits{A.}},
\bauthor{\bsnm{E}, \binits{W.}}:
\batitle{Solving high-dimensional partial differential equations using deep learning}.
\bjtitle{Proceedings of the National Academy of Sciences}
\bvolume{115}(\bissue{34}),
\bfpage{8505}--\blpage{8510}
(\byear{2018})
\end{barticle}
\endbibitem

\bibitem[\protect\citeauthoryear{He et~al.}{2016}]{he2016deep}
\begin{bchapter}
\bauthor{\bsnm{He}, \binits{K.}},
\bauthor{\bsnm{Zhang}, \binits{X.}},
\bauthor{\bsnm{Ren}, \binits{S.}},
\bauthor{\bsnm{Sun}, \binits{J.}}:
\bctitle{Deep residual learning for image recognition}.
In: \bbtitle{Proceedings of the IEEE Conference on Computer Vision and Pattern Recognition},
pp. \bfpage{770}--\blpage{778}
(\byear{2016})
\end{bchapter}
\endbibitem

\bibitem[\protect\citeauthoryear{Bengio et~al.}{1994}]{bengio1994learning}
\begin{barticle}
\bauthor{\bsnm{Bengio}, \binits{Y.}},
\bauthor{\bsnm{Simard}, \binits{P.}},
\bauthor{\bsnm{Frasconi}, \binits{P.}}:
\batitle{Learning long-term dependencies with gradient descent is difficult}.
\bjtitle{IEEE transactions on neural networks}
\bvolume{5}(\bissue{2}),
\bfpage{157}--\blpage{166}
(\byear{1994})
\end{barticle}
\endbibitem

\bibitem[\protect\citeauthoryear{Hochreiter}{1998}]{hochreiter1998vanishing}
\begin{barticle}
\bauthor{\bsnm{Hochreiter}, \binits{S.}}:
\batitle{The vanishing gradient problem during learning recurrent neural nets and problem solutions}.
\bjtitle{International Journal of Uncertainty, Fuzziness and Knowledge-Based Systems}
\bvolume{6}(\bissue{02}),
\bfpage{107}--\blpage{116}
(\byear{1998})
\end{barticle}
\endbibitem

\bibitem[\protect\citeauthoryear{Gazoulis et~al.}{2023}]{gazoulis2023stability}
\begin{botherref}
\oauthor{\bsnm{Gazoulis}, \binits{D.}},
\oauthor{\bsnm{Gkanis}, \binits{I.}},
\oauthor{\bsnm{Makridakis}, \binits{C.G.}}:
On the stability and convergence of physics informed neural networks.
arXiv preprint arXiv:2308.05423
(2023)
\end{botherref}
\endbibitem

\bibitem[\protect\citeauthoryear{Yuan et~al.}{2022}]{yuan2022pinn}
\begin{barticle}
\bauthor{\bsnm{Yuan}, \binits{L.}},
\bauthor{\bsnm{Ni}, \binits{Y.-Q.}},
\bauthor{\bsnm{Deng}, \binits{X.-Y.}},
\bauthor{\bsnm{Hao}, \binits{S.}}:
\batitle{A-pinn: Auxiliary physics informed neural networks for forward and inverse problems of nonlinear integro-differential equations}.
\bjtitle{Journal of Computational Physics}
\bvolume{462},
\bfpage{111260}
(\byear{2022})
\end{barticle}
\endbibitem

\bibitem[\protect\citeauthoryear{McClenny and Braga-Neto}{2023}]{mcclenny2023self}
\begin{barticle}
\bauthor{\bsnm{McClenny}, \binits{L.D.}},
\bauthor{\bsnm{Braga-Neto}, \binits{U.M.}}:
\batitle{Self-adaptive physics-informed neural networks}.
\bjtitle{Journal of Computational Physics}
\bvolume{474},
\bfpage{111722}
(\byear{2023})
\end{barticle}
\endbibitem

\bibitem[\protect\citeauthoryear{Aghaei et~al.}{2024}]{aghaei2024pinniesefficientphysicsinformedneural}
\begin{botherref}
\oauthor{\bsnm{Aghaei}, \binits{A.A.}},
\oauthor{\bsnm{Moghaddam}, \binits{M.M.}},
\oauthor{\bsnm{Parand}, \binits{K.}}:
PINNIES: An Efficient Physics-Informed Neural Network Framework to Integral Operator Problems
(2024).
\url{https://arxiv.org/abs/2409.01899}
\end{botherref}
\endbibitem

\bibitem[\protect\citeauthoryear{Wang et~al.}{2008}]{wang2008finite}
\begin{barticle}
\bauthor{\bsnm{Wang}, \binits{X.}},
\bauthor{\bsnm{Wildman}, \binits{R.A.}},
\bauthor{\bsnm{Weile}, \binits{D.S.}},
\bauthor{\bsnm{Monk}, \binits{P.}}:
\batitle{A finite difference delay modeling approach to the discretization of the time domain integral equations of electromagnetics}.
\bjtitle{IEEE Transactions on Antennas and Propagation}
\bvolume{56}(\bissue{8}),
\bfpage{2442}--\blpage{2452}
(\byear{2008})
\end{barticle}
\endbibitem

\bibitem[\protect\citeauthoryear{Brambilla et~al.}{2008}]{brambilla2008integral}
\begin{barticle}
\bauthor{\bsnm{Brambilla}, \binits{R.}},
\bauthor{\bsnm{Grilli}, \binits{F.}},
\bauthor{\bsnm{Martini}, \binits{L.}},
\bauthor{\bsnm{Sirois}, \binits{F.}}:
\batitle{Integral equations for the current density in thin conductors and their solution by the finite-element method}.
\bjtitle{Superconductor Science and Technology}
\bvolume{21}(\bissue{10}),
\bfpage{105008}
(\byear{2008})
\end{barticle}
\endbibitem

\bibitem[\protect\citeauthoryear{Vandandoo et~al.}{2024}]{vandandoo2024high}
\begin{bbook}
\bauthor{\bsnm{Vandandoo}, \binits{U.}},
\bauthor{\bsnm{Zhanlav}, \binits{T.}},
\bauthor{\bsnm{Chuluunbaatar}, \binits{O.}},
\bauthor{\bsnm{Gusev}, \binits{A.}},
\bauthor{\bsnm{Vinitsky}, \binits{S.}},
\bauthor{\bsnm{Chuluunbaatar}, \binits{G.}}:
\bbtitle{High-Order Finite Difference and Finite Element Methods for Solving Some Partial Differential Equations}.
\bpublisher{Springer}, \blocation{???}
(\byear{2024})
\end{bbook}
\endbibitem

\bibitem[\protect\citeauthoryear{Hamfeldt and Lesniewski}{2021}]{Hamfeldt2021}
\begin{barticle}
\bauthor{\bsnm{Hamfeldt}, \binits{B.F.}},
\bauthor{\bsnm{Lesniewski}, \binits{J.}}:
\batitle{Convergent finite difference methods for fully nonlinear elliptic equations in three dimensions}.
\bjtitle{Journal of Scientific Computing}
\bvolume{90}(\bissue{1}),
\bfpage{35}
(\byear{2021})
\doiurl{10.1007/s10915-021-01714-6}
\end{barticle}
\endbibitem

\bibitem[\protect\citeauthoryear{Sarakorn and Vachiratienchai}{2018}]{Sarakorn2018}
\begin{barticle}
\bauthor{\bsnm{Sarakorn}, \binits{W.}},
\bauthor{\bsnm{Vachiratienchai}, \binits{C.}}:
\batitle{Hybrid finite difference--finite element method to incorporate topography and bathymetry for two-dimensional magnetotelluric modeling}.
\bjtitle{Earth, Planets and Space}
\bvolume{70}(\bissue{1}),
\bfpage{103}
(\byear{2018})
\doiurl{10.1186/s40623-018-0876-7}
\end{barticle}
\endbibitem

\bibitem[\protect\citeauthoryear{Liu et~al.}{2022}]{Liu2022}
\begin{barticle}
\bauthor{\bsnm{Liu}, \binits{W.K.}},
\bauthor{\bsnm{Li}, \binits{S.}},
\bauthor{\bsnm{Park}, \binits{H.S.}}:
\batitle{Eighty years of the finite element method: Birth, evolution, and future}.
\bjtitle{Archives of Computational Methods in Engineering}
\bvolume{29}(\bissue{6}),
\bfpage{4431}--\blpage{4453}
(\byear{2022})
\doiurl{10.1007/s11831-022-09740-9}
\end{barticle}
\endbibitem

\bibitem[\protect\citeauthoryear{Nath et~al.}{2024}]{Nath2024}
\begin{barticle}
\bauthor{\bsnm{Nath}, \binits{D.}},
\bauthor{\bsnm{{Ankit}}},
\bauthor{\bsnm{Neog}, \binits{D.R.}},
\bauthor{\bsnm{Gautam}, \binits{S.S.}}:
\batitle{Application of machine learning and deep learning in finite element analysis: A comprehensive review}.
\bjtitle{Archives of Computational Methods in Engineering}
\bvolume{31}(\bissue{5}),
\bfpage{2945}--\blpage{2984}
(\byear{2024})
\doiurl{10.1007/s11831-024-10063-0}
\end{barticle}
\endbibitem

\bibitem[\protect\citeauthoryear{Chen et~al.}{2023}]{math11214418}
\begin{botherref}
\oauthor{\bsnm{Chen}, \binits{X.}},
\oauthor{\bsnm{Zhang}, \binits{K.}},
\oauthor{\bsnm{Ji}, \binits{Z.}},
\oauthor{\bsnm{Shen}, \binits{X.}},
\oauthor{\bsnm{Liu}, \binits{P.}},
\oauthor{\bsnm{Zhang}, \binits{L.}},
\oauthor{\bsnm{Wang}, \binits{J.}},
\oauthor{\bsnm{Yao}, \binits{J.}}:
Progress and challenges of integrated machine learning and traditional numerical algorithms: Taking reservoir numerical simulation as an example.
Mathematics
\textbf{11}(21)
(2023)
\doiurl{10.3390/math11214418}
\end{botherref}
\endbibitem

\bibitem[\protect\citeauthoryear{Effati and Buzhabadi}{2012}]{effati2012neural}
\begin{barticle}
\bauthor{\bsnm{Effati}, \binits{S.}},
\bauthor{\bsnm{Buzhabadi}, \binits{R.}}:
\batitle{A neural network approach for solving fredholm integral equations of the second kind}.
\bjtitle{Neural Computing and Applications}
\bvolume{21},
\bfpage{843}--\blpage{852}
(\byear{2012})
\end{barticle}
\endbibitem

\bibitem[\protect\citeauthoryear{Mosleh}{2014}]{mosleh2014numerical}
\begin{barticle}
\bauthor{\bsnm{Mosleh}, \binits{M.}}:
\batitle{Numerical solution of fuzzy linear fredholm integro-differential equation by$\backslash$$\backslash$fuzzy neural network}.
\bjtitle{Iranian journal of fuzzy systems}
\bvolume{11}(\bissue{1}),
\bfpage{91}--\blpage{112}
(\byear{2014})
\end{barticle}
\endbibitem

\bibitem[\protect\citeauthoryear{Jafarian et~al.}{2015}]{jafarian2015artificial}
\begin{barticle}
\bauthor{\bsnm{Jafarian}, \binits{A.}},
\bauthor{\bsnm{Measoomy}, \binits{S.}},
\bauthor{\bsnm{Abbasbandy}, \binits{S.}}:
\batitle{Artificial neural networks based modeling for solving volterra integral equations system}.
\bjtitle{Applied Soft Computing}
\bvolume{27},
\bfpage{391}--\blpage{398}
(\byear{2015})
\end{barticle}
\endbibitem

\bibitem[\protect\citeauthoryear{Asady et~al.}{2014}]{asady2014utilizing}
\begin{barticle}
\bauthor{\bsnm{Asady}, \binits{B.}},
\bauthor{\bsnm{Hakimzadegan}, \binits{F.}},
\bauthor{\bsnm{Nazarlue}, \binits{R.}}:
\batitle{Utilizing artificial neural network approach for solving two-dimensional integral equations}.
\bjtitle{Mathematical Sciences}
\bvolume{8},
\bfpage{1}--\blpage{9}
(\byear{2014})
\end{barticle}
\endbibitem

\bibitem[\protect\citeauthoryear{Chaharborj et~al.}{2017}]{chaharborj2017study}
\begin{barticle}
\bauthor{\bsnm{Chaharborj}, \binits{S.S.}},
\bauthor{\bsnm{Chaharborj}, \binits{S.}},
\bauthor{\bsnm{Mahmoudi}, \binits{Y.}}:
\batitle{Study of fractional order integro-differential equations by using chebyshev neural network}.
\bjtitle{J. Math. Stat}
\bvolume{13}(\bissue{1}),
\bfpage{1}--\blpage{13}
(\byear{2017})
\end{barticle}
\endbibitem

\bibitem[\protect\citeauthoryear{Guo et~al.}{2021}]{guo2021physics}
\begin{barticle}
\bauthor{\bsnm{Guo}, \binits{R.}},
\bauthor{\bsnm{Shan}, \binits{T.}},
\bauthor{\bsnm{Song}, \binits{X.}},
\bauthor{\bsnm{Li}, \binits{M.}},
\bauthor{\bsnm{Yang}, \binits{F.}},
\bauthor{\bsnm{Xu}, \binits{S.}},
\bauthor{\bsnm{Abubakar}, \binits{A.}}:
\batitle{Physics embedded deep neural network for solving volume integral equation: 2-d case}.
\bjtitle{IEEE Transactions on Antennas and Propagation}
\bvolume{70}(\bissue{8}),
\bfpage{6135}--\blpage{6147}
(\byear{2021})
\end{barticle}
\endbibitem

\bibitem[\protect\citeauthoryear{Lu et~al.}{2021}]{Lu2021}
\begin{barticle}
\bauthor{\bsnm{Lu}, \binits{L.}},
\bauthor{\bsnm{Jin}, \binits{P.}},
\bauthor{\bsnm{Pang}, \binits{G.}},
\bauthor{\bsnm{Zhang}, \binits{Z.}},
\bauthor{\bsnm{Karniadakis}, \binits{G.E.}}:
\batitle{Learning nonlinear operators via deeponet based on the universal approximation theorem of operators}.
\bjtitle{Nature Machine Intelligence}
\bvolume{3}(\bissue{3}),
\bfpage{218}--\blpage{229}
(\byear{2021})
\doiurl{10.1038/s42256-021-00302-5}
\end{barticle}
\endbibitem

\bibitem[\protect\citeauthoryear{Li et~al.}{2021}]{li2021fourierneuraloperatorparametric}
\begin{botherref}
\oauthor{\bsnm{Li}, \binits{Z.}},
\oauthor{\bsnm{Kovachki}, \binits{N.}},
\oauthor{\bsnm{Azizzadenesheli}, \binits{K.}},
\oauthor{\bsnm{Liu}, \binits{B.}},
\oauthor{\bsnm{Bhattacharya}, \binits{K.}},
\oauthor{\bsnm{Stuart}, \binits{A.}},
\oauthor{\bsnm{Anandkumar}, \binits{A.}}:
Fourier Neural Operator for Parametric Partial Differential Equations
(2021).
\url{https://arxiv.org/abs/2010.08895}
\end{botherref}
\endbibitem

\bibitem[\protect\citeauthoryear{Martire et~al.}{2022}]{martire2022fractional}
\begin{bbook}
\bauthor{\bsnm{Martire}, \binits{A.L.}},
\bauthor{\bsnm{Congedo}, \binits{M.A.}},
\bauthor{\bsnm{Cenci}, \binits{M.}}:
\bbtitle{Fractional Volterra Integral Equations: A Neural Network Approach}
vol. \bseriesno{1}.
\bpublisher{Roma TrE-Press}, \blocation{???}
(\byear{2022})
\end{bbook}
\endbibitem

\bibitem[\protect\citeauthoryear{Saneifard et~al.}{2022}]{Saneifard_Jafarian_Ghalami_Nia_2022}
\begin{barticle}
\bauthor{\bsnm{Saneifard}, \binits{R.}},
\bauthor{\bsnm{Jafarian}, \binits{A.}},
\bauthor{\bsnm{Ghalami}, \binits{N.}},
\bauthor{\bsnm{Nia}, \binits{S.M.}}:
\batitle{Extended artificial neural networks approach for solving two-dimensional fractional-order volterra-type integro-differential equations}.
\bjtitle{Information Sciences}
\bvolume{612},
\bfpage{887}--\blpage{897}
(\byear{2022})
\doiurl{10.1016/j.ins.2022.09.017}
\end{barticle}
\endbibitem

\bibitem[\protect\citeauthoryear{Hornik et~al.}{1990}]{hornik1990universal}
\begin{barticle}
\bauthor{\bsnm{Hornik}, \binits{K.}},
\bauthor{\bsnm{Stinchcombe}, \binits{M.}},
\bauthor{\bsnm{White}, \binits{H.}}:
\batitle{Universal approximation of an unknown mapping and its derivatives using multilayer feedforward networks}.
\bjtitle{Neural networks}
\bvolume{3}(\bissue{5}),
\bfpage{551}--\blpage{560}
(\byear{1990})
\end{barticle}
\endbibitem

\bibitem[\protect\citeauthoryear{Pinkus}{1999}]{pinkus1999approximation}
\begin{barticle}
\bauthor{\bsnm{Pinkus}, \binits{A.}}:
\batitle{Approximation theory of the mlp model in neural networks}.
\bjtitle{Acta numerica}
\bvolume{8},
\bfpage{143}--\blpage{195}
(\byear{1999})
\end{barticle}
\endbibitem

\bibitem[\protect\citeauthoryear{Mhaskar}{1996}]{mhaskar1996neural}
\begin{barticle}
\bauthor{\bsnm{Mhaskar}, \binits{H.N.}}:
\batitle{Neural networks for optimal approximation of smooth and analytic functions}.
\bjtitle{Neural computation}
\bvolume{8}(\bissue{1}),
\bfpage{164}--\blpage{177}
(\byear{1996})
\end{barticle}
\endbibitem

\end{thebibliography}
\end{document}